\documentclass[acmsmall]{acmart}

\AtBeginDocument{%
  \providecommand\BibTeX{{%
    \normalfont B\kern-0.5em{\scshape i\kern-0.25em b}\kern-0.8em\TeX}}}

\usepackage{soul}
\usepackage{multirow}

\begin{document}

\title{Computational Approaches to the Detection of Lesser-Known Rhetorical Figures: A Systematic Survey and Research Challenges}
\author{Ramona Kühn}
\email{ramona.kuehn@uni-passau.de}
\affiliation{%
  \institution{Chair of Data Science, CAROLL Research Group}
  \streetaddress{Innstraße 43}
  \city{Passau}
  \country{Germany}
  \postcode{94032}
}
\author{Jelena Mitrović}
\email{jelena.mitrovic@uni-passau.de}
\affiliation{%
  \institution{Chair of Data Science, CAROLL Research Group}
  \streetaddress{Innstraße 43}
  \city{Passau}
  \country{Germany}
  \postcode{94032}
}
\affiliation{%
  \institution{Institute for AI Research and Development of Serbia}
  \streetaddress{Fru\v{s}kogorska 1 21000 Novi Sad}
  \city{21000 Novi Sad}
  \country{Serbia}
  \postcode{21000}
}
\author{Michael Granitzer}
\email{michael.granitzer@uni-passau.de}
\affiliation{%
  \institution{Chair of Data Science}
  \streetaddress{Innstraße 43}
  \city{Passau}
  \country{Germany}
  \postcode{94032}
}








\renewcommand{\shortauthors}{Kühn et al.}

\begin{abstract}


Rhetorical figures play a major role in our everyday communication as they make text more interesting, more memorable, or more persuasive. Therefore, it is important to computationally detect rhetorical figures to fully understand the meaning of a text. We provide a comprehensive overview of computational approaches to lesser-known rhetorical figures. We explore the linguistic and computational perspectives on rhetorical figures, emphasizing their significance for the domain of Natural Language Processing. We present different figures in detail, delving into datasets, definitions, rhetorical functions, and detection approaches. We identified challenges such as dataset scarcity, language limitations, and reliance on rule-based methods.

\end{abstract}





\maketitle
\section{Introduction}
\label{sec:intro}
Rhetorical figures,\footnote{In the English language, rhetorical figures are often referred to as figures of speech. However, we consider figures of speech only a subcategory of rhetorical figures. The terms rhetorical devices and rhetorical figures are considered the same and are used interchangeably.} also known as rhetorical devices, are language constructions that are used in all forms 
of communication to convey a persuasive message, add emphasis, create vivid imagery, evoke emotions, increase memorability, and engage the audience. This is achieved by repeating words or phrases, figurative language, humor, word plays, exaggeration, or understatement. Examples of rhetorical figures include metaphor, alliteration, irony, or sarcasm.


A clear definition of what rhetorical figures are does not exist. However, most definitions 
describe ``the notion that the figures somehow represent departures from the normal usage'' \cite{fahnestock2002rhetorical}. 
This means that they depart from ``ordinary signifying practices,[...] ordinary syntax'', or from the normal form. Of course, the question arises what ``normal'' usage means. It can be considered as a deviation from a straightforward syntax without repetitions. As every rhetorical figure has a specific function~\cite{crocker1977social,harris2018annotation,harris2023rules}, it is important to fully understand them and grasp their subtle or implicit meanings. Especially in the digital era, where text is generated in vast amounts, processing, analyzing, and understanding it automatically becomes essential. However, modern natural language processing (NLP) tools often fail in tasks such as sentiment analysis or hate speech detection when rhetorical figures come into play. This means that the performance of all of these tasks can be increased when rhetorical figures are detected and understood (see Section~\ref{subsec:compPerspective} for more examples).
However, current approaches only focus on a handful of popular figures, such as metaphor, irony, and sarcasm. Furthermore, as with most research in NLP, the focus lies on the English language.






\subsection{Contributions and Structure of the Paper}
In this survey, we comprehensively explore the landscape of rhetorical figures from a computational perspective. \textbf{Our unique contribution lies in shedding light on computational detection approaches for lesser-known or lesser-investigated figures} such as antithesis, epanalepsis, or zeugma, which are nevertheless prevalent in all languages.

Other surveys in the domain of the computational detection of rhetorical figures mainly focus on metaphors, such as the surveys by Shutova~\cite{shutova2010models}, Rai and Chakraverty~\cite{rai2020survey}, or the survey by Ge et al.~\cite{ge2023survey} demonstrate.
Although Abulaish et al.~\cite{abulaish2020survey} consider in their survey also figures other than metaphor (i.e., hyperbole, simile, metaphor, sarcasm, irony), their focus lies still on popular and well-known figures. Furthermore, they only focus on Twitter as a data source, which can be highly deceptive (see Section~\ref{sec:currentChallenges}).

In contrast to those surveys, our work explicitly investigates figures that are less omnipresent in current research but nevertheless omnipresent in our daily communication. 
We focus on rhetorical figures, their formal descriptions, their definitions and functions, datasets, and different detection techniques. However, we purposefully abstain from considering computational approaches to the interpretation of figures due to the lack of research in this field.
\begin{figure}[h]
    \centering
    \includegraphics[width=1\textwidth]{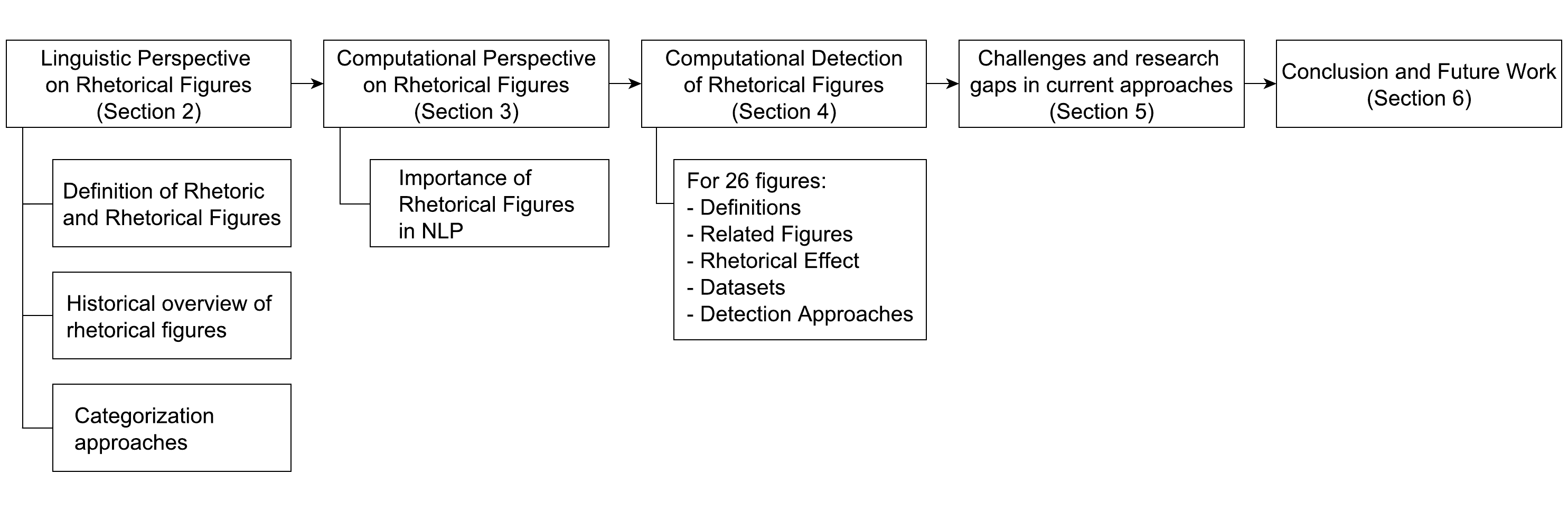}
    \caption{The structure of this survey. We cover the linguistic background briefly. Our main focus lies on the formalization of rhetorical figures, annotated datasets, and the detection of various figures. We are not considering the computational interpretation of rhetorical figures.}
    \label{fig:overview}
    \Description[Structure of the survey]{Section 2 is about the linguistic perspective on rhetorical figures, including definitions of rhetoric and rhetorical figures, a historical overview, and categorization approaches. Section 3 describes how important rhetorical figures are for NLP. Section 4 shows for 24 different figures the definitions, relations, rhetorical effects, datasets, and how they can be computationally detected. Section 5 summarizes our findings and reveals the challenges that we identified. Section 6 ends the survey with a conclusion and suggestions for future work.}
\end{figure}
Fig.~\ref{fig:overview} outlines the structure of our survey. We first present the linguistic view on rhetorical figures and how diverse definitions and spellings of rhetorical figures evolved over time (Section~\ref{sec:lingview}). We then present the computational view on rhetorical figures in Section~\ref{subsec:compPerspective}, highlighting the importance of rhetorical figures in different domains of NLP. In Section~\ref{sec:figureDetection}, we investigate 24 lesser-known rhetorical figures and techniques to computationally detect them. We present authors, their papers about the detection of rhetorical figures, a classification of their approach, a short description, and performance metrics if available. As we can only present static information in this paper, we made additional information available online and each column sortable to better compare the different aspects. The table is available online~\footnote{\url{www.ramonakuehn.de/survey}} From our insights, we compiled in Section~\ref{sec:currentChallenges} a list of the biggest challenges researchers face when computationally detecting rhetorical figures and suggest solutions for each challenge.

In summary, our survey explores the domain of rhetorical figure detection. By combining linguistic insights with computational challenges and solutions, we contribute to a holistic understanding of this topic. Our emphasis on lesser-known figures aims to increase awareness of their importance and the challenges they come along with, guiding future research endeavors and advancements in the field of NLP.

\section{The Linguistic Perspective on Rhetoric and Rhetorical Figures}
\label{sec:lingview}
In this section, we give an overview of the basics of rhetoric, the historical development of rhetorical figures, and the attempts to categorize them over the centuries. 
Aristotle described in his treatise ``Rhetoric'' (approx. 4th century BC) the field of \textbf{rhetoric} as a part of an argument, besides logic and dialectic.
In his definition, rhetoric focuses on how to present arguments, involving emotions and persuasion techniques.

We purposely abstain from providing a universally valid definition of the term ``rhetoric'' here, as it is ``an incredibly difficult word to define because it refers to several different but related concepts''~\cite{covino2014rhetoric}. Kronman~\cite{kronman1998rhetoric} describes ``[r]hetoric is the art of persuading people to believe things.'' As the reader might guess at this point, Kronman also criticizes that this definition is too shallow, not describing how rhetoric can precisely achieve this, and also questioning if the aspect of persuasion is mandatory.
Meyer~\cite{meyer2017rhetoric} attempts to define rhetoric as a  ``negotiation between individuals -- ethos and pathos -- on a question (logos) which divides them to a greater or lesser degree or purports to abolish  or at least diminish their distance.'' However, he also states that the boundaries are too fuzzy, making rhetoric a discipline without well-formed, unified definitions. 

What scholars agree on, however, is that rhetoric can be further divided into the rhetorical appeals of \textbf{logos}, \textbf{ethos}, and \textbf{pathos}. These three form the so-called rhetorical triangle. 
Each of those appeals needs to be addressed to create a persuasive and appealing speech.
Ethos refers to establishing the credibility and authority of the speaker to show the audience that he or she is trustworthy. While rhetorical figures can enhance a speaker's ethos indirectly by showcasing their linguistic and rhetorical skills, they are not inherently ethos-related. Logos appeals to logic and reason. It involves using facts, evidence, and sound reasoning to persuade an audience. Rhetorical figures can be used to strengthen logical arguments or make them more persuasive, but they are not considered to be part of logos. Pathos is an appeal to emotion. It is where rhetorical figures are situated. because they are used to evoke emotions, engage the audience, or add memorability and impact to a message.

It seems that recurrent patterns in rhetoric are the varying, unspecific, and ambiguous definitions. We want to shed light on how those varying definitions could evolve over the centuries, especially for rhetorical figures. One of the first known works, where rhetorical figures are described in detail is the ``Ad Herennium'' (author unknown, approx. 80 BC). The author has already tried to categorize different figures into the two groups of figures of diction and figures of thought. 
However, ten figures from the group of figures of diction are distinguished further, creating actually three categories. The distinction is not self-explanatory. Fahnestock~\cite{fahnestock2002rhetorical} even describes it as ``odd'', creating ``confusion'' as it ``uses no consistent principles of sorting''. Quintilian published his Institutio oratoria~\cite{quintilian1996institutio} approx. 90 AD. He also tried to categorize, classify, and describe rhetorical figures. However, his distinction is also not free from contradictions. Further research that followed, e.g., by St. Augustine, Richard Whately, Ethelbert William Bullinger, and Kenneth Burke, but none of them managed to make a clear separation between different categories of figures.

A rather recent standard literature of rhetorical figures is the book of Lausberg~\cite{lausberg1990handbuch} (1990). It contains examples and explanations in English, French, Greek, Latin, and even more languages. Unfortunately, the mixture of languages, the sometimes contradictory explanations, and the structure make the book difficult to understand. However, Lausberg is aware of this problem. There were so many attempts to establish a single truth in the previous centuries that multiple truths were established, with slightly different definitions that are often too broad or contradict each other. This makes a computational approach to rhetorical figures even more challenging.

Nowadays, there are also digital sources available. One of the most important is the Silva Rhetoricae\footnote{\url{http://rhetoric.byu.edu/}}\cite{burton2007forest}, which provides definitions and examples in English, including sources, pronunciation, related figures, Greek and Latin terms of figures or alternative spellings. In total, it consists of 433 figures. The figures are not categorized but listed in alphabetical order.
Another project is the RhetFig database\footnote{\url{https://artsresearch.uwaterloo.ca/chiastic/display/}} with its origins in the RhetFig project~\cite{kelly2010toward}. It shows a graph of a selected rhetorical figure, where its cognitive affinity (repetition, position), the linguistic domain (e.g., syntactic, lexical), and the taxonomic class (e.g., scheme, trope) of the figure are represented as nodes connected through vertices with the figure itself. 



As time has shown, categorizing rhetorical figures is an infeasible task~\cite{mayer2007rhetorische}.
Burton says that ``sometimes it is difficult to see the forest (the big picture) of rhetoric because of the trees (the hundreds of Greek and Latin terms naming figures of speech, etc.) within rhetoric.''~\cite{burton2007forest}.  Furthermore, some figures are different in other languages or do not have a matching equivalent~\cite{wang2022towards, zhu2022configure,kuhn_2024_10698380}.

In summary, although rhetorical figures are omnipresent and have been studied for centuries, this field is rather chaotic. Nevertheless, they are an essential part of our communication. In the following section ~\ref{subsec:compPerspective}, we will highlight how they improve various NLP tasks and how ontologies can help to bring order to the ambiguous definitions.

\section{The Computational Perspective on Rhetorical Figures}
\label{subsec:compPerspective}
Compared to the linguistic perspective on rhetorical figures, the computational perspective is a rather recent area of research. It is mainly situated in the field of NLP. 
Researchers in this domain aim to enable computers to understand, interpret, and generate human-like language. 


Rhetorical figures and their often non-literal meaning pose a challenge to many NLP systems, as they are often not able to correctly interpret the content of a message. 

Neglecting the existence of rhetorical figures can result in higher error rates and misclassification, particularly in scenarios where implicit language, such as sarcasm or metaphors, is prevalent~\cite{macavaney2019hate}. We will show in the following different NLP tasks that would benefit from the detection of rhetorical figures and demonstrate which tasks have already successfully led to performance improvement.



\textbf{Propaganda, Fake News, and Abusive Language:} Hamilton~\cite{hamilton2021towards} highlights the importance of rhetorical figures for propaganda detection. 
The figure hyperbole is known to be frequently present in fake news~\cite{fang2019self,dwivedi2021survey,rubin2016fake,troiano2018computational} and false advertisements, which can be especially dangerous in the health domain~\cite{troiano2018computational}.
The association between sarcasm and abusive language further highlights the interplay of rhetorical figures in hate speech detection \cite{frenda2022killing}. Notably, Frenda et al. \cite{frenda2023sarcasm} revealed a substantial overlap between tweets classified as hateful and those containing sarcasm, emphasizing the necessity of considering rhetorical nuances for accurate classification.
Lemmens et al.~\cite{lemmens2021improving} were able to improve the performance of their hate speech detection system when accounting for rhetorical figures. Understanding the coexistence of sarcasm and abusive language is essential for developing robust NLP models capable of discerning nuanced expressions.  Zhu et al.~\cite{zhu2021self} mention the necessity of identifying euphemisms to facilitate the work of content moderators in online forums to avoid illegal acts. The figure dysphemism is often used in implicit hate speech~\cite{magu2018determining,felt2020recognizing}.

\textbf{Grammar Detection, Text Summarization, and Information Extraction:} Beyond hate-related tasks, the application of rhetorical figure detection extends to grammatical error identification. For instance, the detection of zeugma has been integrated into a Czech proofreader system to enhance the identification of grammatical errors \cite{medkova2020automatic}. This exemplifies the versatility of rhetorical figure detection in addressing linguistic challenges beyond the realm of semantic analysis. Alliheedi et al.~\cite{alliheedi2014rhetorical} improved the performance of text summarization by considering features of rhetorical figures. Understanding exaggeration like they appear in hyperbole can help improve dialogue systems and information extraction, as it helps better understand human communication~\cite{troiano2018computational}. Strommer~\cite{strommer2011using} uses rhetorical figures for author attribution.

\textbf{Sentiment, Emotion, and Humor Analysis:} 
When texts from social platforms contain irony, sarcasm, or hyperbole, modern sentiment analyzers often fail to grasp the correct sentiment~\cite{davidov2010semi}. Piccirilli and Schulte~\cite{piccirilli2022features} showed that discourse becomes more emotional when metaphors are used. Chen et al. \cite{chen2021jointly} highlight the relevance of irony in sentiment analysis as they discovered that this figure frequently appears in situations labeled with the emotion "disgust."  Ngyuen et al.~\cite{nguyen2015kelabteam} could improve their sentiment analysis when considering rhetorical figures. Gavidia et al.~\cite{gavidia2022cats} found out that the figure euphemism decreases negative or offensive sentiments.  It is also known that negation, and therefore all figures containing negation such as litotes, plays a highly important role in sentiment analysis~\cite{wiegand2010survey,karp2021meiosis}. Ranganath et al.~\cite{ranganath2016identifying} assume that rhetorical questions are used to strengthen or weaken a sentiment.
Furthermore, several rhetorical figures play an important role of humor~\cite{mihalcea2006learning, mihalcea2010computational,morales2017identifying}. For example, detecting the figure alliteration leads to better results in humor identification in the case of Mihalcea and Strapparava~\cite{mihalcea2006learning}.


\textbf{AI and LLM Improvement:} Troiano et al.~\cite{troiano2018computational} state the importance of understanding exaggeration to improve AI systems, as it gives a clearer picture of human communication. This statement can be considered true for any other figure. 
Zhan and Wan~\cite{zhang2021mover} also highlight the importance of rhetorical figures in AI, as generative language models still tend to produce more literal text. Rhetorical figures, however, are a necessary part of authentic human communication. It is important to have larger datasets of rhetorical figures to be able to train and improve machine learning models and LLMs~\cite{dubremetz2015rhetorical, zhang2021mover}.

\section{Computational Detection of Rhetorical Figures}
\label{sec:figureDetection}
This section presents different methodologies to computationally detect rhetorical figures. A manual search was conducted to retrieve relevant papers. We avoided relying on databases like the ACL Anthology. Given the relatively limited attention that rhetorical figures have experienced in bigger NLP conferences, our exploration is extended to publications from smaller conferences. Our strategy involved searching on Google Scholar\footnote{\url{https://scholar.google.com/}} for the figure names along with ``detection'' or ``identification''. We realized that in some cases, the name of a figure is not included in the title of a paper. By considering related work, we uncovered additional relevant research.

For each figure, we offer a general textual \textit{definition} and examples, and discuss associated problems and edge cases. We identified \textit{related figures} that shall help researchers to identify figures that might be detected with similar strategies. The \textit{rhetorical effect} describes the function the figure has on the readers. The \textit{domain} describes areas in which this rhetorical figure seems to appear more frequently. We also list for each figure \textit{datasets} if they are available. In the paragraph \textit{detection}, we present different detection approaches for each figure. If the detection was executed in languages other than English, it is explicitly stated; otherwise, it is assumed to be in English.
We group similar figures together to avoid redundant descriptions that arise when investigating multiple similar figures in one paper. This approach does not only minimize duplication but also highlights hierarchical dependencies among figures (e.g., every figure of perfect lexical repetition, such as anaphora, is also a ploke~\cite{wang2021ontology}).
At the end of each section, we present a table offering a comprehensive overview of the various approaches. We categorized them into two categories.
\begin{itemize}
    \item Rule-based approaches rely on detecting specific word constructions, dictionaries, or lists of words.
    \item Model-based means those approaches use computational or mathematical models, e.g., SVM, LSTM, or BERT.
\end{itemize}
In some cases, authors try both approach directions or a combination. If available, we further indicate the best performance results they achieve. However, it is important to exercise caution when interpreting performance metrics, as variations in datasets and language resources available for the considered language often constrain their comparability. If only precision and recall are given, we calculate and add the F1-score for comparison.

\subsection{Alliteration}
\label{subsec:alliteration}
\paragraph{Definition} The Silva Rhetoricae~\cite{burton2007forest} describes the figure as ``the repetition of the same letter or sound within nearby words. Most often, repeated initial consonants.'' While an alliteration can start with the same first letter, the sound of the letter is crucial. For this reason, it is mostly considered a phonetic figure rather than a syntactic. An example is ``\underline{t}wo \underline{t}errible \underline{t}igers''.

\paragraph{Related Figures} Alliteration can be considered related to figures of repetition, but on a phonetic and not syntactic level.
\paragraph{Rhetorical Effect} Mihalcea and Strapparava~\cite{mihalcea2006learning} discover that one-line jokes contain alliteration, e.g., ``Veni, Vidi, Visa: I came, I saw, I did a little shopping.'' (note: besides alliteration, this sentence also contains parallelism, asyndeton, and anaphora). It calls to the attention of the readers and listeners. Mihalcea et al.~\cite{mihalcea2010computational} describe that it has the effect of inducing expectation.
\paragraph{Domain} Alliteration contributes to humor~\cite{mihalcea2006learning, mihalcea2010computational,morales2017identifying}. It is also a popular figure in poetry, and songs. Often, it is used in combination with an adjective and noun to make the noun more memorable or to mock someone (e.g., Donald Trump called Nancy Pelosi from the other political party ``Nasty Nancy'').

\paragraph{Datasets} No annotated datasets were found for alliteration.
\paragraph{Detection Approach}
Mihalcea and Strapparava~\cite{mihalcea2006learning} aim to identify humorous jokes. They use alliteration as a feature. For its detection, they use the CMU pronunciation dictionary to perform a longest prefix matching. By including alliteration, they achieve better results than when including antonyms. They use the same approach in their later work~\cite{mihalcea2010computational}.

Morales and Zhai~\cite{morales2017identifying} also consider alliteration as a feature in their humor detection task. However, they proceed in the same way as Mihalcea and Strapparava~\cite{mihalcea2006learning} by using the CMU pronunciation dictionary.

\begin{table}[!h]
\begin{tabular}{llllrrr} \toprule
Figure                    & Paper      & Year & Approach & Precision & Recall & F1 \\
\midrule
\multirow{2}{*}{Alliteration}
                           & Mihalcea and Strapparava~\cite{mihalcea2006learning} & 2006 & rule-based &    -     &  -      & -  \\
                          & Morales and Zhai~\cite{morales2017identifying}& 2017 & rule-based &    -     &  -      & -  \\
\bottomrule
\end{tabular}
\caption{Overview of alliteration detection methods. Performance metrics are not available.}
\end{table}

\subsection{Figures with Perfect Lexical Repetition}
\label{subsec:perfectLexicalRepetition}
In this section, we explore rhetorical figures characterized by the syntactic repetition of a word or group of words. The discussed figures include ploke, anadiplosis, anaphora, epiphora, symploke, epanalepsis, epizeuxis, diacope, and conduplicatio. Antimetabole is intentionally excluded from this discussion because it has a broad research scope and is often combined with chiasmus. Therefore, we treat it separately in Section~\ref{subsec:antimetabole}.

\paragraph{Definitions} \textbf{Ploke} (also known as ploce, ploche, or, according to Fahnestock~\cite{fahnestock2002rhetorical} also called heratio, duplicatio, palilogia, diaphora, and  traductio) is the general term for any form of repetition. It describes the perfect lexical repetition of a word or multiple words without consideration of the location of the repetition. Therefore, the detection of any figure involving perfect lexical repetition (e.g., those mentioned in this section) implies the presence of a ploke. However, there can exist instances where word repetition may occur non-rhetorically for emphasizing reasons, such as it is the case with separable verbs. Those verbs do not occur in the English language, but extensively in other languages such as German. For example, in the sentence ``Wir fingen an, an danach zu denken'', ``an'' refers to two distinct concepts. The classification as a rhetorical ploke versus an unintended repetition may be debatable.

An \textbf{anadiplosis} represents a more specific form of ploke, involving the repetition of a word or group of words at the end of one sentence and at the beginning of the next sentence, line, or clause. For instance, "I am \underline{strong}. \underline{Strong} in the face of adversity."

A \textbf{conduplicatio} can be considered as a simple repetition of a word or multiple words. It is related to anadiplosis, yet distinguishes itself by repeating a keyword from the preceding sentence at the beginning of the subsequent sentence.

An \textbf{anaphora} (or epanaphora) is characterized by the repetition of a word or group of words at the commencement of successive sentences, clauses, paragraphs, or lines. It is debatable whether the recurrence of stopwords instigates an anaphora, a consideration often addressed by the approaches discussed below, wherein stopwords are frequently excluded. Additionally, the interpretive distance between sentences remains a subject of discussion. An illustrative instance of an anaphora is in Martin Luther King's renowned speech: ``I have a dream that [....]. I have a dream that [....].''

In contrast, an \textbf{epiphora} (or epistrophe) manifests when consecutive sentences, clauses, or phrases conclude with identical words. For instance, ``We are born to sorrow, pass our time in sorrow, end our days in sorrow''~\cite{burton2007forest}.

When both an anaphora and an epiphora occur in sequential sentences, the resulting figure is termed \textbf{symploke} (or symploce). The co-location of these repetitions in a symploke implies the concurrent presence of an anaphora and an epiphora. Therefore, methods designed for detecting symploke can effectively identify anaphora and epiphora, and vice versa. Bill Clinton's statement is an example: "When there is talk of hatred, let us stand up and talk against it. When there is talk of violence, let us stand up and talk against it."

An \textbf{epanalepsis} involves the repetition of the initial word(s) at the end of a line, phrase, or clause. While the Silva Rhetoricae~\cite{burton2007forest} allows for the repetition of the same word after intermediate words, this definition appears to be relatively uncommon. A salient example of epanalepsis is: "Believe not all you can hear, tell not all you believe."

A \textbf{diacope} describes the repetition of a word with one or more in between.

An \textbf{epizeuxis} describes an immediate, perfect lexical repetition of words. Punctuation in between, such as commas, are allowed. Example: ``Never, never, never''. Its definition differs slightly between languages. In English, the number of repetitions is not further specified, whereas in German the figure epizeuxis often refers to a repetition of three or more times~\cite{wang2022towards}.

\paragraph{Related Figures} All figures that use syntactic repetition are related. Antimetabole also belongs to the figures above, as it also uses repeating words. Polyptoton uses the same words but in a different inflectional form. Alliteration can be seen as a repetition of the same sound.

\paragraph{Rhetorical Effect} Figures with a perfect lexical repetition create rhythm, increase memorability, emphasize the repeated element, and make it appear more significant~\cite{fahnestock2002rhetorical}. Besides the known effects of figures of repetition, epizeuxis creates additional vehemence~\cite{burton2007forest}, while an epanalepsis conveys a feeling of circularity and completeness of an argument.
\paragraph{Domain}
Figures with perfect lexical repetition can appear in all domains. Dubremetz and Nivre~\cite{dubremetz2018rhetorical} experience that anaphoras appear more often in the domain of fiction than in the domain of science.
Music is also a huge domain where repetition plays a crucial role. Songs with repeating rhythms and lyrics often become favorite songs and people listen to them again and again, while a repetition in texts quickly becomes boring~\cite{margulis2014repeat}. A frequent pattern in music is ploke. In some cases, the part to repeat is indicated by a special sign (a straight line and two dots, or if only one bar is repeated, a slash with two dots). Another frequent pattern is epanalepsis. It is indicated with the so-called Da Capo al fine sign (``from the beginning to the end''), indicating that at the end of the musical piece, the musician jumps back to the beginning to repeat the first parts until a \textit{Fine} sign indicates the end.

\paragraph{Datasets}
Both Gawryjolek~\cite{gawryjolek2009automated} and 
Java~\cite{java2015characterization} collect a few examples of anaphora, epiphora, anadiplosis, antimetabole, epanalepsis, epizeuxis, and ploke. However, the dataset is not publicly accessible. Another problem is that Gawryjolek only uses positive instances in the dataset. Both datasets encompass only a small number (between 25 to maximum 50) of instances. 
Dubremetz and Nivre annotate anaphoras and epiphoras in their original antimetabole dataset~\cite{dubremetz2018rhetorical}. 

Zhu et al.~\cite{zhu2022configure} create a Chinese corpus for contextualized figure recognition (ConFiguRe), which they built from 98 Chinese literary works such as novels and prose. 
The corpus contains besides other figures around 450 instances of repetition. As they do not further specify the location of repetition, it is comparable to ploke.

\paragraph{Detection}
Gawryjolek~\cite{gawryjolek2009automated} is one of the first documenting how to computationally detect rhetorical figures. He focuses on the figures anadiplosis, anaphora, antimetabole (see Section~\ref{subsec:antimetabole}), epanalepsis, epiphora, epizeuxis, ploke, polyptoton (see Section~\ref{subsec:polyptoton}), polysyndeton (see Section~\ref{subsec:polysyndeton}), parallelsim/isocolon (see Section~\ref{subsec:parallelism}), and oxymoron (see Section~\ref{subsec:oxymoron}). His overall goal is to implement an annotation tool for rhetorical figures and visualize their usage in political speeches. 
He tests his approach on collected examples from political speeches, Wikipedia, Silva Rhetoricae, American Rhetoric, about.com, Bible and many more. He collects between 20 and 52 examples for each figure, however, he does not specify the exact number or how many of them are correctly identified by his algorithms. A further issue we remark is that he has only positive examples in his dataset, but mentions the performance of precision on his approach, which cannot be calculated without having false positives. What he is actually referring to is recall. In his evaluation, he focuses rather on examples that were not correctly identified, but he does not provide specific numbers.
For \textbf{ploke}, he ignores stopwords. Furthermore, he specifies boundaries in which the repetition has to occur. For the detection of \textbf{anadiplosis}, he checks if a group of words at the end of a syntactic unit is also repeated at the beginning of the next unit. He does not consider conjunctions and determiners. 

For \textbf{anaphora}, Gawryjolek considers both repeating words and group of repeating words at the beginning. He first searches for a repetition of words and then determines the location to find an anaphora. He set a minimum phrase length but realized that there are also two-word phrases that can contain an anaphora, therefore, he changed the minimum phrase length to two.

For the detection of \textbf{epiphora}, Gawryjolek proceeds similarly to the detection of anaphora by first finding repeating words and then determining their location. However, he does not try the detection of symploke which is a compound figure, consisting of an anaphora and epiphora combined. To find \textbf{epanalepsis} Gawryjolek looks for a repeating group of words at the beginning of a sentence and then searches if the same group is repeated at the end of the same syntactic unit. Gawryjolek looks for an immediate perfect repetition of words to detect an \textbf{epizeuxis}.

Java~\cite{java2015characterization} extends Gawryjolek's approaches. He claims that he can identify 14 rhetorical figures, especially figures of repetition such as 
anadiplosis, anaphora, antimetabole, conduplicatio, epanalepsis, epiphora, epizeuxis, ploke, polyptoton (see Section~\ref{subsec:polyptoton}), polysyndeton (see Section~\ref{subsec:polysyndeton}), and symploke. Further, he tries to identify isocolon (see Section~\ref{subsec:parallelism}), chiasmus (see Section~\ref{subsec:antimetabole}), and oxymoron (see Section~\ref{subsec:oxymoron}).
For figures of repetition he uses a search window whose size varies for each figure (e.g., one sentence, or two sentences). Within this window, he is looking for repeating words.
As a dataset, he creates a file with at least 25 instances of the rhetorical figure under investigation. The examples are collected from the Bible, not further specified literature, political speeches, popular culture, common sayings, and rhetorical websites. For \textbf{anadiplosis}, he has a total number of 42, where only one was not detected, resulting in a precision of 100~\% and a recall of 97.6~\%. He does not state if the data contains negative examples, and if so, how many. Furthermore, the dataset is rather small and not representative. Dubremetz and Nivre~\cite{dubremetz2017machine} mention as one of the challenges in the detection of rhetorical figures the differentiation between salient repetitions and accidental repetitions. We think that Java's approach is not able to make this distinction. He uses the detection of rhetorical figures for authorship attribution, but he does not clarify why he focuses on the mentioned figures.

For \textbf{ploke}, Java searches for a repetition of words within two sentences, where he ignores stopwords. He achieves a precision and recall of 100~\%.

For \textbf{epanalepsis}, Java again ignores leading determiners, conjunctions, and prepositions. The search window spans over one single sentence. On the 29 instances, Java achieves a precision of 97~\% and a recall of 100~\%.

Java uses again his search window enlarged to three sentences to identify \textbf{anaphoras} in his dataset containing 29 examples of this figure. He achieves both a precision and recall of 100~\%. He ignores conjunctions, prepositions, and leading determines. However, this leads to the problem that multiple repeating words at the beginning of consecutive sentences starting e.g., with a ``The'' will be ignored. Problems with this method were revealed by Dubremetz and Nivre~\cite{dubremetz2018rhetorical} three years later that leads to sentences starting with ``The'' are ignored while being salient examples of anaphoras.
To detect \textbf{epiphora}, Java~\cite{java2015characterization} sets the search window to three sentences, looking for repeating words at the end, allowing also sentences not to be consecutive. On the 42 instances, he has a precision of 95~\% and a recall of 100~\%.
For \textbf{epizeuxis}, Java looks within one sentence for immediate repetitions of a word or a phrase where the case is ignored. He achieves both a precision and recall of 100~\% on his 42 instances.
Java also considers the detection of \textbf{conduplicatio}, achieving a precision and recall of 100~\% on his 25 instances. Its detection is similar to the one of Ploke, but refers to repeating phrases. However, Java does not provide further information how those phrases are defined. Nevertheless, he achieves a precision and recall of 100~\% on his 25 collected instances.
We want to remark that the performance metrics of Java have to be taken with a grain of salt. He often achieves high precision and recall. However, the dataset is very small and it is likely that he has only very salient instances of the mentioned figures in the data.

Hromada~\cite{hromada2011initial} uses PERL-compatible regular expressions to detect an \textbf{anadiplosis} amongst other rhetorical figures such as \textbf{anaphora}, \textbf{antimetabole} (see Section~\ref{subsec:antimetabole}), and \textbf{epiphora} based on the repetition of words. The expressions are applied to different texts in English, German, Latin, and French to prove the language independence of their approach. Unfortunately, performance metrics are not reported as the corpora is not annotated.

For \textbf{anaphora} Strommer~\cite{strommer2011using} uses decision trees, SVMs, logistic regression, bayesian classification, and adaptive boosting to classify anaphoras. The approach where he achieves the highest F1-score is naive Bayes combined with sentence length.

Lawrence et al.~\cite{lawrence2017harnessing} present a small study on the figures anadiplosis, epanaphora, epistrophe, epizeuxis, eutrepismus, polyptoton, dirimens copulatio, and antithesis. As a dataset, they used 12 episodes from the summer seasons of BBC's Moral Maze in 2012. For the detection of \textbf{anadiplosis}, they split text in constitutive dialogue and associated propositional units and then simply compare words with each other.
For each figure, they suggest a different rule-based approach, often looking for specific words. However, their approach appears to work only on very salient and classic examples of rhetorical figures. Furthermore, they do not report common performance metrics such as recall or precision, and their division of text into clauses is solely based on punctuation marks.
For anaphora, Lawrence et al. filter out clauses that begin with the same words.
The same approach is used to detect epiphoras, only that they filter out sentences that end with the same words.
For the figure \textbf{epizeuxis}, Lawrence et al. consider uni-, bi-, and tri-grams to find at least two repetitions.

Dubremetz and Nivre~\cite{dubremetz2018rhetorical} investigate the detection of \textbf{anaphora}, \textbf{epiphora}, and chiasmus. Similar to their previous work on chiasmus~\cite{dubremetz2015rhetorical}, they consider anaphora and epiphora detection not as a binary classification but rather as a ranking task for salient and not salient examples. They consider successive sentences with one identical initial lemma. Sentences beginning with ``The'' or other non-meaningful words are not considered. This causes the problem that consecutive sentences starting with ``The'' and the same words afterward as the previous sentence are not considered an anaphora while being a very salient one. With different features (e.g., punctuation, identical bigrams, size difference, etc.), they test their approach. Including all features leads to the best result in regards to F1-score.

Lagutina et al.~\cite{lagutina2020automatic} also investigate figures containing a lexical repetition for the sake of investigating the change of rhythm in literature over centuries. They consider the investigated figures as rhythmical because of their lexical repetition. They look into the detection of anadiplosis, anaphora, polysyndeton~\ref{subsec:polysyndeton}, 
diacope, epanalepsis, epiphora, epizeuxis, and symploke. For anadiplosis, anaphora, epiphora, they reuse the algorithm from their previous paper~\cite{lagutina2019automated}, which is written in Russian. They do not report any performance metrics of their detection algorithm, as they are not using an annotated dataset.
In their recent work, they use 150 fiction novels from known English and Russian authors as data. The experiments are conducted in English and Russian. For anadiplosis, they search for a repeating word or a repeating pair of words that are separated by one punctuation mark. For \textbf{epanalepsis}, Lagutina et al.~\cite{lagutina2020automatic} look if a word repeated in the first half of the sentence also ends the sentence. For \textbf{anaphora}, they collect first a list of neighboring sentences and then check if they start with the same word. For \textbf{epiphora}, they proceed similarly by searching for the repetition at the end of a sentence. They extend their searches as long as the word is different. For their symploke detection, they simply combine their anaphora and epiphora algorithms.
Diacope is described by Lagutina et al.~\cite{lagutina2020automatic} as the ``repetition of a word or phrase with intervening words within one sentence''. For its detection, they search for unique words and collect those that are not adjacent/next to each other. A word has to be repeated at least two times to form a diacope.
To find \textbf{epizeuxis}, Lagutina et al. go through the sentences and compare if the neighboring sentence repeats the same sentence or if a word in the first half of the sentence is repeated two times.

Zhu et al.~\cite{zhu2022configure} also investigate the Chinese figure \textbf{repetition} (mostly resembles the English ploke), besides many others. They first define figurative units as the smallest continuous clause sequence carrying a complete expression of a specific figure. A clause is separated by punctuation. For the detection of repetition, they use E2ESeq model and E2ECRF with additional CRF layer. Unfortunately, they do not report performance metrics for individual figures. They only report the performance of the combination of all twelve figures they are investigating.
\begin{table}[!h]
\begin{tabular}{llllrrr} \toprule
Figure                    & Paper      & Year & Approach & Precision & Recall & F1 \\
\midrule
\multirow{3}{*}{Ploke} 
                        
                        & Gawryjolek~\cite{gawryjolek2009automated} & 2011 & rule-based  &      -   &    -    & -  \\
                        & Java~\cite{java2015characterization}      & 2015 &   rule-based        &  100.00  & 100.00 & \textbf{100.00} \\  
                        & Zhu et al.~\cite{zhu2022configure} & 2022 & model-based& 31.10	& 31.10 & 31.10 \\                        \midrule

\multirow{8}{*}{Anaphora} 
                        & Gawryjolek~\cite{gawryjolek2009automated} & 2011 & rule-based  &     -    &    -    &  - \\
                        & Strommer~\cite{strommer2011using}    & 2011 &    model-based       &   90.90     &  89.90 & 86.10  \\
                        & Hromada~\cite{hromada2011initial}    & 2011 &    rule-based       &    -    & - & - \\
                        & Java~\cite{java2015characterization}      & 2015 &   rule-based        &   100.00     &  100.00 & \textbf{100.00} \\ 
                        & Lawrence et al.~\cite{lawrence2017harnessing}            &   2017   &     rule-based       &   -     & -  & -\\    
                        & Dubremetz and Nivre~\cite{dubremetz2018rhetorical}& 2018 & model-based & -&- &63.73
\\
                        & Lagutina et al.~\cite{lagutina2020automatic}   & 2020 &  rule-based         & -&    -    &  -  \\ \midrule
\multirow{6}{*}{Epiphora} 
                        & Gawryjolek~\cite{gawryjolek2009automated} & 2011 &  rule-based          &    -    &  -  & -\\
                        & Hromada~\cite{hromada2011initial}    & 2011 &    rule-based        &     -   &  -  & - \\
                        & Java~\cite{java2015characterization}       & 2015 &   rule-based         &   95.00     &  100   & \textbf{97.44} \\
                        & Lawrence et al.~\cite{lawrence2017harnessing}            &     2017       &   rule-based  &    -   & -  & - \\    
                        & Dubremetz and Nivre~\cite{dubremetz2018rhetorical}& 2018 & model-based & -&- &51.80
\\
                        & Lagutina et al.~\cite{lagutina2020automatic} &    2020  &    rule-based       &    -    &  -   &- \\  \midrule
\multirow{2}{*}{Symploke} 
                        & Java~\cite{java2015characterization}      & 2015 &   rule-based        &  100.00  & 97.00 & \textbf{98.48} \\
                        & Lagutina et al.~\cite{lagutina2020automatic} & 2020 &        rule-based    &   -     & -   & -\\    \midrule
\multirow{3}{*}{Anadiplosis} 
                        & Gawryjolek~\cite{gawryjolek2009automated} & 2011 &   rule-based         &  -      &   - & - \\
                        & Lawrence et al.~\cite{lawrence2017harnessing}            &    2017  &      rule-based     &   -   & -  & - \\
                        & Lagutina et al.~\cite{lagutina2020automatic} & 2020 &     rule-based       & -       &  -  &- \\                         \midrule
\multirow{1}{*}{Conduplicatio} 
                        & Java~\cite{java2015characterization}      & 2015 &   rule-based        &   100.00 & 100.00 & \textbf{100.00} \\ \midrule
\multirow{2}{*}{Epanalepsis} 
                        & Java~\cite{java2015characterization}      & 2015 &   rule-based        &   97.00 & 100.00 & \textbf{98.48} \\
                        & Lagutina et al.~\cite{lagutina2020automatic} & 2020 &      rule-based      &    -    &  -  & -\\    \midrule
\multirow{1}{*}{Diacope} 
                        & Lagutina et al.~\cite{lagutina2020automatic} & 2020 &      rule-based      &   -     &  -  &- \\\midrule
\multirow{4}{*}{Epizeuxis} 
                        & Gawryjolek~\cite{gawryjolek2009automated} & 2011 &    rule-based        &  -      & -   & -\\
                        & Java~\cite{java2015characterization}      & 2015 &   rule-based        & 100.00   & 100.00 &  \textbf{100.00}\\
                        & Lawrence et al.~\cite{lawrence2017harnessing}            &    2017  &    rule-based       &  -      & -  & -\\  
                        & Lagutina et al.~\cite{lagutina2020automatic} & 2020 &       rule-based     &   -     &  -  &- \\ 
\bottomrule
\end{tabular}
\caption{Overview of the performance of computational detection approaches for figures of perfect lexical repetition.}
\end{table}


\subsection{Antimetabole and Chiasmus}
\label{subsec:antimetabole}
\label{subsec:chiasmus}
The antimetabole is a more specific form of the figure chiasmus. Therefore, we encounter several papers that claim to look into chiasmus but actually investigate the figure antimetabole, where the same words are syntactically repeated in a criss-cross reversed order. In contrast, chiasmus can use words that are only semantically related but not syntactically similar.

\paragraph{Definitions} \textbf{Chiasmus} is the reversal of semantically equivalent words or concepts/ideas. It is often confused with antimetabole, where syntactically equivalent words are repeated in reverse order. According to the Silva Rhetoricae, it can even be the reverse repetition of a grammatical structure~\cite{burton2007forest}. An example of a chiasmus is ``By day the frolic, and the dance by night'' (Samuel Johnson).
\textbf{Antimetabole} repeats the same words in reverse order, following a criss-cross pattern. It is a more specific form of chiasmus, where semantic equal words are repeated in reverse order. Example: ``For I do not seek to \underline{understand} in order to \underline{believe}, but I \underline{believe} in order to \underline{understand}''

\paragraph{Related Figures} Parallelism and antithesis can co-occur with antimetabole~\cite{green2020towards}. Antimetabole is also related to figures with perfect lexical repetition such as anadiplosis, epiphora, asyndeton, antimetabole, epanalepsis, epizeuxis, ploke, and polysyndeton. However, an antimetabole can also repeat words in different inflectional forms.

\paragraph{Rhetorical Effect} Antimetaboles create rhythm and balance. The reversed repetition often highlights a contrast and call to action, making the message more impactful. Chiasmus can create symmetry but also highlights contrast.

\paragraph{Domain} Dubremetz and Nivre~\cite{dubremetz2018rhetorical} state that antimetabole and chiasmus appear more often in scientific than fictional contexts.

\paragraph{Datasets}
Gawryjolek~\cite{gawryjolek2009automated} collects instances of antimetabole but does not specify how many. Java~\cite{java2015characterization} collects 25 instances of antimetabole and 33 instances of chiasmus.
Dubremetz and Nivre~\cite{dubremetz2015rhetorical} create an antimetabole dataset, although they refer to it as a chiasmus dataset. They collect examples from English fiction books, scientific articles, and examples from quotes from websites. It is available online.\footnote{\url{https://github.com/mardub1635/corpus-rhetoric}}
Schneider et al.~\cite{Schneider23:HIT} build a German chiasmus/antimetabole dataset, but we could not find it online.

\paragraph{Detection}

The first approach to detect antimetabole is performed by Gawryjolek~\cite{gawryjolek2009automated}. He looks for words repeating exactly in reverse order, calling it ``word palindrom''. However, this also causes the detection of constructions that are actually not salient to be an antimetabole and produces a lot of false positives. Furthermore, he looks for exact repetition, so antimetaboles with changed inflections or word forms are not detected.

This approach is extended by Hromada~\cite{hromada2011initial} with PERL-compatible regular expressions to detect antimetabole.

Java~\cite{java2015characterization} considers for the antimetabole detection only words that are either nouns, verbs, adjectives, or adverbs. However, he only focuses on the repetition within one sentence and not over multiple ones. On the dataset with 25 antimetaboles, he achieves a precision of 71~\% and a recall of 100~\%. However, as the data is not available to us, we assume that the dataset does not contain any edge cases where the antimetabole is arguable or non-salient, making this a rather basic approach only for obvious antimetaboles.
Chiasmus is considered a ``reversal of grammatical structure'' by Java. However, this differs from most definitions that see chiasmus as a reversal of words with the same meaning or synonyms, focusing more on a semantic than syntactic aspect. Therefore, Java only considers the reversal of POS tag equivalence classes, where he groups POS tags into the classes adjective, noun, adverb, verb, and pronoun. Still, he achieves only a precision of 50~\% and a recall of 42.4~\% on his 33 instances of what he considers chiasmus.

Dubremetz, one of the most prominent researchers for antimetabole and chiasmus, started to investigate the different variants of how antimetaboles can appear~\cite{dubremetz2013towards}. 
In the earlier work of Dubremetz and Nivre~\cite{dubremetz2015rhetorical}, they call the figure chiasmus but actually focus on the detection of antimetabole, which is a specific form of chiasmus. Other research in the domain of antimetaboles and chiasmus also criticizes that Dubremetz and Nivre mainly look into antimetabole in a broader sense and not chiasmus~\cite{harrisantimetabole, schneider2021data}.
An important contribution of Dubremetz is that they identify that detection of antimetabole (or chiasmus respectively) is not binary but rather a ranking task.
They include features such as removing stopwords and punctuation, size-related features, n-gram features, and lexical clues. They also describe a ranking function. They test their approach on the English Europarl dataset. They annotate 1200 criss-cross patterns to be either a chiasmus (or antimetabole), not a chiasmus, or borderline. However, in 200 top hits of their algorithm, they only encounter 19 antimetaboles. Their average precision is  61~\%, the recall is around the same. But with increasing recall, the precision decreases drastically. They achieve the best metrics with the approach with lexical cues.
The further development of this approach is presented in their following work~\cite{dubremetz2016syntax}. They also include features such as POS tagging and syntactical dependency structures. They achieve a precision of 67~\% on their test set. As a point of critique, they annotate only the top 200 patterns their algorithm retrieved. This also means that they are completely missing examples that the algorithm did not find at all. They apply their algorithm to a new dataset that contains 200 candidates and only 8 real antimetaboles. The baseline finds 6 of them, whereas their model finds 7 antimetaboles. However, the authors remark that the results are not significant on this small dataset. They test an SVM with oversampling and logistic regression on their English dataset from previous research~\cite{dubremetz2015rhetorical, dubremetz2016syntax}. The logistic regression approach works best when including all features (e.g., POS tags, syntax dependencies, etc.). It results in a precision of 90~\%, a recall of 69.2~\%, and an F1 of 78.3~\%.
The later work of Dubremetz and Nivre~\cite{dubremetz2018rhetorical} discusses antimetabole again as a special form of chiasmus, but does not present new results for this figure. The work focuses on applying the approach to epiphora and epanaphora (see Section~\ref{subsec:perfectLexicalRepetition}).

Schneider et al.~\cite{schneider2021data} investigate chiasmus as a repetition of semantically similar words, but also test their approach on Dubremetz's antimetabole dataset. They first filter suitable chiasmus candidates by relying on POS tags instead of lemmas. The candidate phrase length is limited to 30 tokens. Then, they filter the candidates with logistic regression. They use Dubremetz's~\cite{dubremetz2017machine} features, lexical features, and embedding features. They train their classifier on four annotated dramas of Friedrich Schiller in German. They test their approach on unseen dramas of Friedrich Schiller, of which they annotate the top 100 candidates. For antimetabole detection on Dubremetz's dataset, they achieve a precision of 73~\% when using Dubremetz's features combined with embedding features. Considering lexical features additionally does not improve the performance further. However, the precision is still lower than Dubremetz's. Schneider et al. do not provide further performance metrics for a better comparison. On the German dataset, Schneider et al. achieve a precision of 49~\% for antimetabole detection when using Dubremetz's and lexical features.

Schneider et al.~\cite{Schneider23:HIT} recently claim that they achieve a precision of 74~\% on their newly created German chiasmus antimetabole dataset. Unfortunately, they do not specify whether it accounts for antimetabole or chiasmus detection. Furthermore, no full text of this research is available.

Meyer~\cite{meyer2023application} considers both Dubremetz's and Schneider's approach~\cite{schneider2021data}, but decides to ignore Schneider's because of ``conceptual flaws'' of the algorithm based on POS-tags. Meyer also uses a sliding window over 30 tokens. As a dataset, he uses Dubremetz's examples and extracts further antimetaboles from a book called ``Never let a fool kiss you, or a kiss fool you''. He tests different approaches (logistic regression, SVM, random forests, regression trees) and includes four additional features to the ones of Dubremetz and Schneider. Logistic regression achieves the best results with 77,5~\% precision and 94.6~\% recall. Meyer is aware of the distinction between antimetabole and chiasmus. He also extends his antimetabole approach to the detection of chiasmus. However, he mentions that it needs further refinement as it cannot detect all chiasmi.

Berthomet~\cite{berthomet2023detecting} has a similar strategy to solve the antimetabole detection task as Meyer, but focuses more on deep learning instead of classical machine learning algorithms. Unlike Dubremetz, he considers antimetabole detection as a classification problem and not as a ranking task. He uses GPTNeo\footnote{\url{https://huggingface.co/docs/transformers/model_doc/gpt_neo}} to detect both antimetabole and chiasmus (see Section~\ref{subsec:chiasmus}, but also remarks that the approach performs better for antimetaboles. In conclusion, Berthomet considers the lack of word position encoding in the transformers a problem for the detection of antimetabole and chiasmus, as the location of words plays a major role. As an improvement, he suggests using a ranking system to encode location in transformers and to encode features into transformer models, as they only take a series of raw text.


\begin{table}[!h]
\begin{tabular}{llllrrr} \toprule
Figure                    & Paper      & Year & Approach & Precision & Recall & F1 \\
\midrule
\multirow{10}{*}{Antimetabole}
                           & Gawryjolek~\cite{gawryjolek2009automated} & 2009 & rule-based &    -     &  -      & -  \\
                          & Hromada~\cite{hromada2011initial}    & 2011 &    rule-based       &  -      & -  & - \\
                          & Java~\cite{java2015characterization}      & 2015 &   rule-based        &   71.00     &  100.00 & 83.04 \\
                          &  Dubremetz and Nivre~\cite{dubremetz2015rhetorical}& 2015 & rule-based & 61.00& - & -\\
                          & Dubremetz and Nivre~\cite{dubremetz2016syntax}& 2016 & rule-based & 67.65 & - & -\\
                          & Dubremetz and Nivre~\cite{dubremetz2017machine}& 2017 & model-based &90.00 & 69.20 &  78.30\\
                          &  Dubremetz and Nivre~\cite{dubremetz2018rhetorical}& 2018 & model-based &90.00 & 69.20 &  78.30\\
                         & Schneider et al.~\cite{schneider2021data}        & 2021     &      model-based     &     73.00   & -    & -\\
                        &    Berthomet~\cite{berthomet2023detecting}        &   2023   &  model-based         &    98.00	& 98.00	&\textbf{98.00} \\
                          & Meyer~\cite{meyer2023application}  &   2023   &     model-based      &    77.50	& 94.60     & 85.20 \\ 

                           & Schneider et al.~\cite{Schneider23:HIT}  &    2023  &       -    &  74.00      & -    & -\\ \midrule
\multirow{5}{*}{Chiasmus} & Gawryjolek~\cite{gawryjolek2009automated} & 2009 &    rule-based       &        -&   - &- \\
                          & Hromada~\cite{hromada2011initial}    & 2011 &     rule-based      &   -     & -   & - \\
                          & Java~\cite{java2015characterization}      & 2015 &    rule-based       &  50.00 &	42.40     & \textbf{45.89} \\
                          & Schneider et al.~\cite{schneider2021data}        & 2021     &      model-based     &     28.00   & -    & - \\
                          & Schneider et al.~\cite{Schneider23:HIT}  &    2023  &       -    &  74.00      & -    & -\\

\bottomrule
\end{tabular}
\caption{Overview of antimetabole and chiasmus detection methods and their performance.}
\end{table}

\subsection{Antithesis}
\label{subsec:antithesis}

\paragraph{Definition} Antithesis uses antonyms in two (often syntactic parallel) phrases or sentences. For some rhetoricians, a syntactic parallel structure is not necessarily required~\cite{burton2007forest,mcguigan2011rhetorical}, whereas others consider it rather mandatory~\cite{fahnestock2002rhetorical}. A very strong, double antithesis is ``working all day, sleeping all night.''

\paragraph{Rhetorical Effect} Antithesis is often used in arguments to compare two sides, where the speaker or writer favors one part over the other. It balances contrasting ideas and creates rhythm when used with a syntactic parallelism.

\paragraph{Related Figures} Parallelism is often part of an antithesis. Oxymoron is related to antithesis as it also contains antonymic relations between words. Antimetabole and chiasmus are also related, as the antimetabole dataset of Dubremetz and Nivre~\cite{dubremetz2015rhetorical} contains several antitheses~\cite{green2020towards}.

\paragraph{Domain} Green~\cite{green2021some} investigates the use of antithesis in environmental science policy articles, where it is used to juxtapose the two different views to preserve nature vs. it can be adapted to human needs. Kühn et al.~\cite{kuhn2023hidden} assume that antithesis occurs more frequently in populistic texts.

\paragraph{Datasets}
Zhu et al.~\cite{zhu2022configure} have around 250 antitheses in their Chinese corpus ConFiguRe. Green and Crotts~\cite{green2020towards} use the antimetabole corpus of Dubremetz and Nivre~\cite{dubremetz2015rhetorical} to extract 120 antitheses from it. They also indicate the occurrence of parison. However, the dataset is not publicly available. Kühn et al.~\cite{kuhn2023hidden} use data from Telegram chats of a German journalist criticizing the government during the Covid-19 pandemic. They extract parallel phrases and annotate in their dataset the occurrences of antitheses. The dataset is in German.

\paragraph{Detection}

Lawrence et al.~\cite{lawrence2017harnessing} approach antithesis detection that the presence of antonyms suffices, not taking into account parallelism or negations such as ``is'' vs. ``is not''. They use WordNet\footnote{\url{https://wordnet.princeton.edu/}} to find antonyms.

Green and Crotts~\cite{green2020towards} use a broader definition of antonyms than Lawrence et al.~\cite{lawrence2017harnessing}, as they also allow synonyms of antonyms. They use a rule-based approach, looking for antonyms in a list they generated from WordNet and ConceptNet\footnote{\url{https://conceptnet.io/}}. As dataset, they use the antimetabole dataset created by Dubremetz and annotate antitheses in it. The authors are at the same time the annotators, and the first ones to create guidelines for the antithesis annotation task. They define three different types of antitheses.

Zhu et al.~\cite{zhu2022configure} define figurative units which represent the smallest continuous clause sequence containing a figure. Punctuation marks decide where a clause ends and begins. For the detection, they use E2ESeq models and E2ECRF with an additional CRF layer. Unfortunately, they do not report individual performance metrics, only the aggregated metrics over all figures they are investigating.

Kühn et al.~\cite{kuhn2023hidden} present a rule-based approach for the detection of antithesis in the German language. As the data from German Wordnets are not satisfactory, they compiled a list of antonyms from the German Wiktionary. The dataset consists of Telegram posts of a COVID-19 skeptic. They hope to find antitheses in there as the skeptic is a populist, trying to show the contracting behavior of the German government at that time. Kühn et al. split the posts into phrases based on punctuation marks and cue words, and search for syntactic parallelism by comparing spaCy POS-tags. If the phrases are parallel, they are looking for a pair of antonyms in their antonym list. However, they only report performance metrics on the antithesis detection task, not on the parallelism detection task.
In their more recent work, Kühn et al.~\cite{kuhn2024using} replace the antonym lists by language models for German. They keep the rule-based parallelism detection in their pipeline. For the detection of antonyms and contrasting ideas, they use the German pre-trained ELECTRA (GELECTRA) language model and achieve good results. They also try different augmentation techniques for the data such as backtranslation and synonym replacement. In addition, they compare the results of GELECTRA with other models such as mBERT and German BERT. Nevertheless, GELECTRA achieves the best results.

\begin{table}[!h]
\begin{tabular}{llllrrr} \toprule
Figure                    & Paper      & Year & Approach & Precision & Recall & F1 \\
\midrule
\multirow{5}{*}{Antithesis} 
                           & Lawrence et al.~\cite{lawrence2017harnessing} & 2017 & rule-based &     -    &   -     & -  \\
                          & Green and Crotts~\cite{green2020towards}    & 2020 &   rule-based       &   41.10 &	38.40 & 39.70 \\
                          & Zhu et al.~\cite{zhu2022configure}    & 2022 &    model-based      & 31.10	&31.10& 		30.99  \\
                          & Kühn et al.~\cite{kuhn2023hidden}      & 2023 &   rule-based        &   57.00 & 45.24& 50.44 \\ 
                          & Kühn et al.~\cite{kuhn2024using} & 2024 & rule-\& model-based & 	83.16	&54.02	&\textbf{65.11} \\

\bottomrule
\end{tabular}
\caption{Overview of approaches to detect antithesis and their performance metrics on different datasets.}
\end{table}

\subsection{Dysphemism and Euphemism}
\label{subsec:dysphemism}
\label{subsec:euphemism}
We will treat dysphemism and euphemism together in one section, as they are often referred to as x-phemism and are highly related.
\paragraph{Definition} \textbf{Dysphemism} uses harsh, offensive, or derogatory language to describe or refer to something. An example of dysphemism is the sentence ``he is worm food now'' to describe someone who died. \textbf{Euphemism} is the opposite figure, which uses milder or more positive language to soften a term or concept, often to make it socially or politically more acceptable. An example of euphemism is saying ``he passed away'' for someone who died, as it sounds softener and more calming.

\paragraph{Related Figures} Euphemism can be related to litotes, when it is used with a double negation or when the opposite is negated. Example: ``This is not good'', when in fact, the situation is really bad.

\paragraph{Rhetorical Effect} With dysphemism, criticism can be expressed, but it can also create a humorous effect, or express strong disdain. Euphemisms are used to soften an argument, often for politeness or to decrease the emotional impact.

\paragraph{Domain} Gavidia et al.~\cite{gavidia2022cats} identify that euphemisms decrease negative or offensive sentiments. When euphemistic terms are replaced by their literal meaning, the sentiment is also more negative or offensive. Furthermore, they state that the figure is often used in the context of polite communication, such as in a respectful dialogue between a doctor and a patient when talking about sensitive medical issues~\cite{rababah2014translatability}.

Euphemisms or dysphemisms can also be used to avoid censoring in online forums by using other words to circumscribe hurtful expressions or illegal activities or objects~\cite{zhu2021self}.
Magu and Luo~\cite{magu2018determining} also identify that euphemisms or codewords are used in online implicit hate speech in Tweets.
Felt and Riloff~\cite{felt2020recognizing} state that euphemisms are used in polite social interactions, whereas dysphemism plays a role in hate speech and abusive language. They also mention that euphemisms are used in politics to justify decisions, and dysphemisms are used in counter-arguments.
Zhu et al.~\cite{zhu2021self} understand euphemism as a figure that appears especially in underground markets and slang language where it is used to secretly circumscribe illegal products or deeds.

\paragraph{Datasets}
Felt and Riloff~\cite{felt2020recognizing} construct a list of x-phemistic terms. With the Basilisk algorithm, they extend this list iteratively by similar terms. Three annotators determine if those terms are dysphemistic, euphemistic, or neutral. A corpus with euphemistic terms in English is built by Gavidia et al.~\cite{gavidia2022cats}. Their corpus is based on the GloWbE Corpus~\cite{davies2015expanding} and examples from online sources. 
Graduate students annotate the corpus with a binary classification scheme (euphemism or not). Furthermore, they indicate their confidence level from 1-3, and explain their decision. The corpus is available online\footnote{\url{https://github.com/marsgav/euphemism_project}}. In the inter-annotator agreement, they reach a rather low Krippendorf alpha with 0.415 compared to annotation in other domains. This highlights the problem that annotators can often not agree on what a rhetorical figure is. 
Adewumi et al.~\cite{adewumi2021potential} have 2384 euphemisms in their dataset of idiomatic expressions.

\paragraph{Detection}

Magu and Luo~\cite{magu2018determining} try to find euphemisms or code words in hateful tweets.
They describe euphemism as an ``instrument to mask intent''. They define a list of code words that are used to circumscribe other concepts (e.g., ``A leppo'' for a Conservative). They use word2vec and calculate cosine distance to find codewords, as they should be more distant from hate speech words. They do not report any performance metrics. In their definition, euphemism is not a way to express something in a softened way but to completely disguise the actual meaning.

Felt and Riloff~\cite{felt2020recognizing} consider the topics fire (as in losing a job), steal, and lying. They use their dataset of annotated x-phemistic terms. A combination of different lexicons leads to better results. They also link dysphemism with high arousal and euphemism with low arousal. Nevertheless, they state that the built lists of x-phemistic terms can never be complete, as new word combinations are constantly developed, and their meaning can change from being a euphemistic term to being an insult and becoming a dysphemistic term.

Zhu et al.~\cite{zhu2021self} investigate the figure euphemism. They also mention euphemism identification, which refers to identifying the literal meaning of the euphemism. They define euphemism as ``ordinary-sounding words with a secret meaning''. They point to further work treating euphemism. However, most of them seem to be rather related to gang, slang, or underground language. 
Zhu et al. use an unsupervised/self-supervised learning approach and focus explicitly on the context of words within a sentence. They treat euphemism detection as a fill-in-the-mask problem that can be solved with BERT in combination with a self-supervised algorithm. They filter out uninformative masked sentences where the masked word is ambiguous. They test different approaches (Word2Vec, TF-idf etc.) on three datasets. Their approach achieves the best results with a precision of 50~\%.

\begin{table}[!ht]
\begin{tabular}{llllrrr} \toprule
Figure                    & Paper      & Year & Approach & Precision & Recall & F1 \\
\midrule
\multirow{3}{*}{Euphemism} 
                          & Magu and Luo~\cite{magu2018determining}  & 2018 & model-based &     -    &   -     & -  \\
                          &Felt and Riloff~\cite{felt2020recognizing}& 2020 & rule-based & - & - & \textbf{67.00} \\
 
                          & Zhu et al.~\cite{zhu2022configure}    & 2022 &   model-based      & 31.10	&31.10& 		30.99  \\
                        \midrule

\multirow{1}{*}{Dysphemism}  & Felt and Riloff~\cite{felt2020recognizing}& 2020 & rule-based & - & - & \textbf{52.00} \\
\bottomrule
\end{tabular}
\caption{Overview of approaches to detect x-phemisms and their performance metrics on different datasets.}
\end{table}

\subsection{Hyperbole}
\label{subsec:hyperbole}
\paragraph{Definition}Hyperbole is characterized by exaggerated and extravagant statements or claims that are often not as true as they might at first. Example: ``The whole world is shocked about this new product!''

\paragraph{Related Figures}Meiosis can be considered as the opposite of a hyperbole. Furthermore, hyperbole is often used with the figure exclamation, where an exclamation mark is used. Further related figures are metaphor, metalepsis, litotes, and similes according to the Silva Rhetoricae~\cite{burton2007forest}.
\paragraph{Rhetorical Effect} Hyperbole is used for emphasis, humor, dramatic effects, and often to increase interest in a topic as it creates a vivid and dramatic effect in language.
\paragraph{Domain}Fake news tends to use the rhetorical figure hyperbole~\cite{fang2019self,dwivedi2021survey,rubin2016fake,troiano2018computational} as an exaggeration in a way that normal news seems to be more interesting. Hyperbole misleads users and can give them a wrong perception of reality. Hyperbole is also used in satire~\cite{rubin2016fake}. Troiano~\cite{troiano2018computational} mention the importance of hyperbole in false advertisement in the medical domain or in exaggerated political statements.
\paragraph{Datasets}
Troiano et al.~\cite{troiano2018computational} build a dataset of exaggeration, called HYPO, and claim that it is the first of its kind, as studies in NLP mainly focus on non-literal figures, but less on figures such as hyperbole. The dataset is created by manually crawling the web and using crowdsource workers to annotate it. It contains 709 instances of hyperboles and also their literal translation (paraphrase). They advise the workers to annotate if the text contains a hyperbole, to mark the hyperbolic word, to rewrite it without an exaggeration, to decide if the expression can be replaced by a number (instead of ``a million times''), and ask them to rank the degree of exaggeration. Their hypothesis is that higher exaggeration leads to a better detection. However, especially the last points led to a big disagreement between the annotators. Troiano et al. also highlight the importance of context for hyperbole and give the example ``It took ages to build the castle''. It is a literal meaning for a real castle made of stone, but a hyperbole when two kids play with sand castles on the beach.
Zhan and Wan~\cite{zhang2021mover} extend this HYPO dataset by training a BERT-based binary classifier on this dataset. They also extract potential hyperboles from online sources. They then label a subset of the retrieved data and retrain their model. They repeat the process for a better hyperbole detection and call this dataset HYPO-XL. It contains 17,862 hyperbolic sentences which adds up to a total of 2.3~\% of the whole dataset. Unlike Troiano et al.~\cite{troiano2018computational}, they do not include the literal translation of the hyperboles.
Zhu et al.~\cite{zhu2022configure} have 690 hyperboles in their Chinese ConFiguRe corpus which is built from 98 Chinese literary works such as novels and prose. 
Adewumi et al.~\cite{adewumi2021potential} create a dataset with idiomatic expressions, where hyperbole is one of them. Their dataset contains 48 hyperboles.

\paragraph{Detection}
Troiano et al.~\cite{troiano2018computational} train different models such as Logistic Regression, Naive Bayes, k-Nearest Neighbor and more on their own crafted dataset. They achieve the best results with Logistic Regression, skip-gram and their features, such as emotional intensity, unexpectedness, polarity, and more.

Zhang and Wan~\cite{zhang2021mover} use a masking approach with BART, a denoising autoencoder
to pretrain sequence-to-sequence models~\cite{lewis2019bart}. Zhang and Wan's assumption is that a single word or phrase triggers hyperbole. They identify spans where the hyperbolic expression is located by unexpectedness calculated by cosine distance and POS-n-grams. Furthermore, they rank their hyperboles according to the strength of the hyperbolic expression. With a RoBERTa base model, they calculate the cosine distance between the literal and the hyperbolic sentence. Only the candidates with the highest score are selected. However, the whole paper focuses more on the generation of hyperbolic expressions than on their detection. Nevertheless, we are confident that the fine-tuned models can be used for hyperbole detection as well. Related to Zhan and Wan's work is Tian et al.~\cite{tian2021hypogen}. They also investigate the generation of hyperboles, but on a sentence level. They collect hyperbole examples from Reddit. However, they only focus on sentences with a ``so ... that'' structure. They also fine-tune a BERT model for binary classification. Next, they apply a ranking scheme to find the most salient examples.


\begin{table}[!h]
\begin{tabular}{llllrrr} \toprule
Figure                    & Paper      & Year & Approach & Precision & Recall & F1 \\
\midrule
\multirow{3}{*}{Hyperbole} 
                          & Troiano et al.~\cite{troiano2018computational}  & 2018 & model-based  &76.00	& 76.00	&\textbf{76.00}  \\
                          & Zhang and Wan~\cite{zhang2021mover} & 2021      &   model-based & - & - &- \\
                          & Tian et al.~\cite{tian2021hypogen} & 2021 & model-based  & 31.00	& 29.00	& 30.00 \\

\bottomrule
\end{tabular}
\caption{Overview of approaches to detect or generate hyperbole.}
\end{table}



\subsection{Irony and Sarcasm}
\label{subsec:irony}
Irony and Sarcasm are popular figures in the realm of NLP. We do not want to investigate them further here as elaborated surveys about them already exist. We want to point the interested reader to the survey of irony from Wallace~\cite{wallace2015computational} and to the survey about sarcasm from Joshi et al.~\cite{joshi2017automatic}.

\subsection{Litotes}
\label{subsec:litotes}
\paragraph{Definition} Litotes uses double negation or the negation of the opposite to express the opposite. However, as Mitrović et al.~\cite{mitrovic2020cognitive} argues, it does not necessarily express the opposite, but a middle position (``the unexcluded middle''). A frequent form or litotes in English is a negation cue, and a negated adjective. It is also often used as a form of understatement. An example is when a person says, ``This is not uninteresting''. The person does not necessarily mean that something is of high interest, it is just more interesting than not interesting at all. This also supports the theory of the unexcluded middle.
\paragraph{Related Figures} Litotes can be related to euphemisms to soften the expression, for example, when a doctor says to the patient ``this is not good'' instead of scaring the patient by directly telling him that he has a severe illness. The concept of producing understatement with litotes is also related to the figure meiosis. It is also often confused with meiosis~\cite{yuan2017argumentative}. Litotes can have an ironic effect. Therefore, it is related to irony~\cite{neuhaus2016relation}.
\paragraph{Rhetorical Effect}In its form of understatement, litotes can often have a humorous aspect, but also disguise the real meaning if the speaker or writer does not want to be too direct. The double negation adds complexity to the sentence, which can confuse readers or listeners.

\paragraph{Domain} Negation in general plays an important role in sentiment analysis~\cite{wiegand2010survey}. Litotes, in particular, has a major effect on sentiment analysis~\cite{karp2021meiosis}. The authors detect that sentences that have a rather positive sentiment but use double negation (``he is not too bad'') are classified as having a negative sentiment. One could only remark that they test this only with one sentiment analyzer and do not make a thorough comparison of different tools for this task.

\paragraph{Datasets}
Mukherjee et al.~\cite{mukherjee2017negait} collect articles about diseases on Wikipedia and had annotators annotate different kinds of negations. Yuan~\cite{yuan2017argumentative} collect double negation in ``The Analects of Confucius''. The dataset contains almost 100 instances of litotes. Paida~\cite{paida2019double} extends the dataset of Yuan by adding more instances to it. The dataset contains 1360 instances of double negation.

\paragraph{Detection}
There is a lot of research about negation, but often it is only focused on simple negation. Double negation has not received that much attention yet.
NegAIT~\cite{mukherjee2017negait} is a system based on a dictionary that contains negation words. The authors try to detect double negation with a rule-based approach.

Paida~\cite{paida2019double} approaches double negation in his master thesis by testing different approaches on his created dataset. He tests a dictionary rule-based approach, SVM, logistic regression, FNN, RNN, and BERT, where BERT outperforms the other approaches.

Karp et al.~\cite{karp2021meiosis} investigate both meiosis and litotes. For litotes, they search with regular expressions in the language R for the occurrence of the words ``not'' and ``bad''. Other words in between are allowed (e.g., ``not too bad''). They investigate the novel ``The Catcher in the Rye'' by Jerome David Salinger, where they also find several instances of litotes. However, they do not report any performance metrics. It is furthermore unclear how many other litotes or meiosis remain hidden due to the lack of manual annotation and a basic approach that considers only one specific construction of litotes.

\begin{table}[!h]
\begin{tabular}{llllrrr} \toprule
Figure                    & Paper      & Year & Approach & Precision & Recall & F1 \\
\midrule
\multirow{3}{*}{Litotes} 
                          & Mukherjee et al.~\cite{mukherjee2017negait}  & 2017 & rule-based & 66.67 & 66.67& 66.67  \\
                          & Paida~\cite{paida2019double} & 2019      &   rule-/model-based &96.00	& 96.00	& \textbf{96.00} \\
                          & Karp et al.~\cite{karp2021meiosis} & 2021 & rule-based & - & - & -  \\

\bottomrule
\end{tabular}
\caption{Overview of approaches to detect litotes and their performance metrics on different datasets.}
\end{table}






\subsection{Meiosis}
\label{subsec:meiosis}
\paragraph{Definition} Meiosis uses diminishing terms or understatement. A meiosis is, for example, saying ``it's just a flesh wound'' (Monty Python and the Holy Grail) when speaking of amputated legs and arms.
\paragraph{Related Figures} Meiosis is considered the opposite of hyperbole. It is also related to litotes and irony.
\paragraph{Rhetorical Effect} Meiosis often creates humor. 
\paragraph{Domain}Like litotes, meiosis has an effect on sentiment analysis~\cite{karp2021meiosis}. Also, it can be used in a humorous way,
\paragraph{Datasets} No annotated datasets of meiosis were encountered.
\paragraph{Detection}

Karp et al.~\cite{karp2021meiosis} investigate both meiosis and litotes, but clearly state that litotes is easier to detect. With the language R, they apply rules to the novel ``The Catcher in the Rye'' by Jerome David Salinger. The authors do not report any performance metrics. This seems to be the only work trying to computationally detect meiosis. However, we wanted to include it in this survey, as it is related to hyperbole.

\begin{table}[!h]
\begin{tabular}{llllrrr} \toprule
Figure                    & Paper      & Year & Approach & Precision & Recall & F1 \\
\midrule
{Meiosis} 
                          & Karp et al.~\cite{karp2021meiosis}  & 2021 & rule-based & - & -& -  \\

\bottomrule
\end{tabular}
\caption{The only approach we found to detect meiosis.}
\end{table}

\subsection{Metaphor}
\label{subsec:metaphor}
Metaphor is the most popular figure in terms of research and computational detection. As we especially focus on less common figures, we will not consider it further here. We refer to the surveys of Rai et al.~\cite{rai2020survey}, Tong et al.~\cite{tong2021recent} and Ge et al.~\cite{ge2023survey}.

\subsection{Metonymy}
\label{subsec:metonymy}
\paragraph{Definition}In metonymy, one word or phrase is substituted with another closely associated word or phrase, typically based on a relationship of contiguity or proximity. Example: ``reading Goethe'' instead of reading literature written by Goethe, or ``Washington said that [...]'' instead of referring to representatives of the government in Washington.
\paragraph{Related Figures} Metaphor is related to metonymy.
\paragraph{Rhetorical Effect} Metonymy often enhances the vividness or symbolism of the expression while conveying a concept like metaphor.
\paragraph{Domain}Metonymy is important for many NLP tasks such as machine translation, geographical information retrieval, entity linking, or coreference resolution~\cite{li2020target}.
\paragraph{Datasets}
Zhu et al.~\cite{zhu2022configure} have in their Chinese ConFiguRe corpus 603 examples of metonymy.

A metonymy dataset was released for the SemEval 2007 Shared Task 8~\cite{markert2007semeval}.
The RELOCAR~\cite{gritta2017vancouver} metonymy dataset is based on that, but tries to overcome the imbalance problem the original SemEval data has~\cite{markert2007semeval}.

A metonymy dataset in German is available online\footnote{\url{https://www.ims.uni-stuttgart.de/forschung/ressourcen/lexika/glmd/}} and was created by Zarcone et al.~\cite{zarcone2012logical}.

WIMCOR~\cite{mathews2020large} is a dataset focusing on location-based metonymies that is based on Wikipedia.

\paragraph{Detection}



Often, metonymy detection focuses only on locations or organizations~\cite{mathews2021impact}. Often, metonymy detection is also referred to as metonymy resolution. Although metonymy is related to metaphor, it did not receive the same attention~\cite{mathews2021impact}. However, our research still found more research on metonymy than for most of the other figures. We, therefore, assume that the detection of metonymy can benefit from approaches to metaphor detection and vice versa. Fass~\cite{fass1991met} combines both figures and presents the met* approach to discriminate between metaphor and metonymy. They present a flow chart in which they investigate metaphorical and anomalous relations. However, the author does not indicate the dataset he used or performance metrics.

Teraoka et al.~\cite{teraoka2012automatic} perform metonymy detection in Japanese. They include associative information between words besides syntactic and semantic information for metonymy detection. They construct associative concept dictionaries for verbs and nouns. On their own test set with 90 sentences, where 45 contain a metonymy, they achieve a precision of 0.85, a recall of 0.73, and an F1-score of 0.79.
 
Gritta et al.~\cite{gritta2017vancouver} consider metonymy as a classification task and introduce a new method called predicate window (PreWin) to combine it with an LSTM. They achieve the best results With an ensemble method. However, they focus only on geographical metonymy. The authors also created a new annotated English dataset for metonymy detection called RELOCAR.

Li et al.~\cite{li2020target} want to overcome the need for hand-crafted dictionaries and lexicon approaches. They use the language models BERT-base and BERT-large both in basic form and also fine-tune them. They test them on various datasets containing spatial metonymies/locations and organizational metonymies. With the BERT-large model fine-tuned on masking, they achieve the best results.

Mathews et al.~\cite{mathews2021impact} find out that the focus on target words does not lead to a big improvement in metonymy detection. However, including context words improves their detection approach. They consider metonymy detection as a sequence labeling task. It is partially based on the one of Gritta et al.~\cite{gritta2017vancouver}. They use the pre-trained BERT base uncased in two variants, where the vector representation of the token is given at each timestep and another variant where the vector representations of the context words of the current token are used to predict the label of the token. Their BERT models perform better than GLOVE. However, the variants do not make a big difference on their WIMCOR dataset.
The authors also present results based on coarse-grained (metonymic or literal) and fine-grained (metonymy related to team, institute, artifact, or event). For the coarse-grained, the two variants do not make a big difference. However, for the fine-grained part, the second variant yields better results.

Wang et al.~\cite{wang2023empirical} extend the approach of Gritta et al.~\cite{gritta2017vancouver} and include more semantic and syntactic features. They test it on the SEMEVAL, RElOCAR, and the WIMCOR dataset. They achieve the best results with an entity-aware BERT model combined with ACGN. On RELOCAR, they have an F1 of 95.7~\% and an accuracy of 95.8~\%.

\begin{table}[!h]
\begin{tabular}{llllrrr} \toprule
Figure                    & Paper      & Year & Approach & Precision & Recall & F1 \\
\midrule
\multirow{5}{*}{Metonymy} 
                          & Fass~\cite{fass1991met}& 1991 & rule-based & - & - & - \\
                          & Teraoka et al.~\cite{teraoka2012automatic}  & 2012 & rule-based & 85.00 & 73.00 & 79.00  \\
                         & Li et al.~\cite{li2020target}& 2020 & model-based & 80.90 & 81.30 & 81.10 \\
                          &Mathews et al.~\cite{mathews2021impact} & 2021& model-based &60.00 & 79.00 & 68.20 \\
                          & Wang et al.~\cite{wang2023empirical}& 2023& model-based & - & - & \textbf{95.80} \\
                                
\bottomrule
\end{tabular}
\caption{Overview of approaches to detect metonymy on different datasets.}
\end{table}

\subsection{Oxymoron}
\label{subsec:oxymoron}
\paragraph{Definition} The Silva Rhetoricae describes oxymorons as ``placing two ordinarily opposing terms adjacent to one another''~\cite{burton2007forest}. Often, it is a noun together with an opposing adjective~\cite{Berner2011}. Following this definition, which represents the majority of rhetoricians, it is not allowed to have words in between the contradicting pair. 
\paragraph{Related Figures} Antithesis is related to oxymoron, as it also uses antonyms and contradiction.

\paragraph{Rhetorical Effect}The juxtaposition of contradictory or opposing words to create a thought-provoking, paradoxical, or vivid expression. It draws attention and conveys emotion, making the word pair more memorable.
\paragraph{Domain} We found no specific domain where oxymorons are more common than in others.
\paragraph{Datasets}
Both Gawryjolek~\cite{gawryjolek2009automated} and Java~\cite{java2015characterization} collect instances of oxymoron. Java has 49 instances in the dataset.
Adewumi et al.~\cite{adewumi2021potential} have 48 examples of oxymorons in their dataset of idiomatic expressions.


La Pietra and Masini~\cite{la2020oxymorons} build an oxymoron dataset in Italian. They use the English antonym list (noun - noun pairs) created by Jones\cite{jones2003antonymy} and translate it to Italian. The lemmas of those pairs are used to search Italian corpora for oxymorons, creating the new dataset. In addition, the authors add the antonymic adverb or adjective to the nouns of their list. Overall, they collect 376 oxymorons.

A dataset of English oxymorons with context-based interpretation (OCBI) is built by Xu et al.~\cite{xu2023creative}.
The authors mention that generating oxymorons by combining antonymic words is hard, as only certain words combined actually make sense. We assume that this assumption is also true for the translation of oxymoron datasets.
Xu et al. collect oxymoron phrases and their context from a website to construct the dataset. They only include sentences that are at least two words longer than the oxymoron. Redundant sentences are removed by applying k-means to combine similar sentences.
\paragraph{Detection}

Gawryjolek~\cite{gawryjolek2009automated} constructs so-called governors how words appear together. He builds pairs of words and uses Wordnet to find if they are antonymous, or if they are synonyms of antonyms, or other derived forms.

Java~\cite{java2015characterization} uses the same approach as Gawryjolek by using Wordnet to look for antonyms, synonyms, and derived forms. He achieves for his 49 examples of oxymorons a precision of 94~\% and a recall of 33~\% with his detection method.

Cho et al.~\cite{cho2017detecting} search for the presence of antonyms by using word vectors and calculating the minimum cosine distance between two verbs, nouns, adjectives, and adverbs. In addition, dependency parse trees are used to verify that the words refer to the same object. They test their approach on English examples. However, they do not mention how the data was annotated. They claim that their approach can be faster in real-world applications than rule-based ones that need to look up antonym pairs in dictionaries. What is noteworthy is that they do not look for antonyms in consecutive words but in the whole phrase. This rather resembles the figure antithesis~\ref{subsec:antithesis}, depending on the definition.


\begin{table}[!h]
\begin{tabular}{llllrrr} \toprule
Figure                    & Paper      & Year & Approach & Precision & Recall & F1 \\
\midrule
\multirow{3}{*}{Oxymoron} 
                          & Gawryjolek~\cite{gawryjolek2009automated}& 2009 & rule-based &-  &-  & - \\
                          & Java~\cite{java2015characterization}  & 2012 & rule-based  & 94.00& 33.00& 48.85 \\
                         & Cho et al.~\cite{cho2017detecting}& 2020 & model-based  & 54.00	& 55.00 & \textbf{55.00} \\
                                
\bottomrule
\end{tabular}
\caption{Overview of approaches to detect oxymoron and their performance metrics on different datasets.}
\end{table}

\subsection{Parallelism/Parison and Isocolon}
\label{subsec:parallelism}
\paragraph{Definition} \textbf{Parallelism}, also called parison, is the repetition of words that belong to the same part of speech category. Example: ``The stronger leads, the weaker follows'' where a determiner, a noun, and a verb are repeated in the same order. The Silva Rhetoricae defines \textbf{isocolon} as ``a series of similarly structured elements having the same length. A kind of parallelism.''~\cite{burton2007forest}. Gawryjolek~\cite{gawryjolek2009automated} also uses this definition. However, Berner~\cite{Berner2011} defines it as words with almost the same amount of syllables in similarly ordered sentences or phrases. Java~\cite{java2015characterization} also considers isocolon as the repetition of the same POS tags, what rather resembles the figure parallelism. There seems to be no clear distinction, so we will consider both figures jointly.

\paragraph{Related Figures} Antithesis is related to parallelism, as it often consists of a parallel structure with antonyms. Antimetaboles can also have a parallel sentence structure.
\paragraph{Rhetorical Effect} Parallelism is known to create balance, rhythm, and emphasize the expression. When each phrase or sentence has the exact same number of syllables, isocolon creates rhythm and increases memorability.
\paragraph{Domain} Parallelism is a very frequent figure across all domains. Therefore, it is encountered in advertisements, political speeches (especially for comparison), and many more.
\paragraph{Datasets}
Both Gawryjolek~\cite{gawryjolek2009automated} and Java~\cite{java2015characterization} consider the detection of isocolon and collect instances of this figure (62 instances in the case of Java).
Chen et al.~\cite{chen2021jointly} collect rhetorical figures in Chinese from literature, textbooks, microblogs, and websites. They have three students annotate rhetorical figures, similes/metaphors, parallelism, personification, rhetorical questions, and irony. In addition, they annotate emotions and show which figure tends to be connected to which emotion.

Adewumi et al.~\cite{adewumi2021potential} have in their dataset with idiomatic expressions 64 instances of parallelism.
Zhu et al.~\cite{zhu2022configure} have 431 examples (called discourse fragments by them) in their Chinese corpus that are considered parallel. Kühn et al.~\cite{kuhn2023hidden} collect and annotate antitheses from an online messenger service. As they extract only parallel phrases, it can be considered to be a parallelism dataset, too.

\paragraph{Detection}
Gawryjolek~\cite{gawryjolek2009automated} describes parallelism but does not further specify its detection. It is subsumed under the detection of isocolon. He considers isocolon as one of the most complicated figures he tried to detect. He looks for repeating POS tags or a maximum deviation of one POS tag. However, this approach does not consider the number of syllables.

Although Java~\cite{java2015characterization} does not explicitly describe the detection of parallelism as his understanding of isocolon resembles more the figure parallelism. We, therefore, think that this approach can also be applied to parallelism. He combines POS tags into equivalence classes, where he groups POS tags referring to all kinds of, e.g., a noun, together.

Kühn et al.~\cite{kuhn2023hidden} uses a similar approach. To find antitheses (a combination of parallelism with antonyms), they first search for sentences containing parallelism by considering repeating POS tags. For the detection, they split the text into smaller units, so-called ``phrases'', at punctuation marks, or certain keywords. Furthermore, they try to achieve parallelism by removing stopwords, quotation marks, and both, as this can lead to a different splitting of phrases. In addition, they apply a Levensthein distance of 75~\% of repeating POS tags to not only focus on strict parallelism. This threshold is based on observations but not on extensive evaluations. Therefore, they do not provide any performance metrics for the detection of parallelism.


\begin{table}[!h]
\begin{tabular}{llllrrr} \toprule
Figure                    & Paper      & Year & Approach & Precision & Recall & F1 \\
\midrule
\multirow{2}{*}{Parallelism} 
& Java~\cite{java2015characterization} & 2015 & rule-based& - & - &  \\
& Kühn et al.~\cite{kuhn2023hidden} & 2023 & rule-based& - & - &  \\

                        \midrule

\multirow{1}{*}{Isocolon}  & Gawryjolek~\cite{gawryjolek2009automated}& 2009 & rule-based & - & - &  \\
& Java~\cite{java2015characterization} & 2015 & rule-based& - & - &  \\
\bottomrule
\end{tabular}
\caption{Overview of approaches to detect parallelism and isocolon. Performance metrics are not available.}
\end{table}

\subsection{Polyptoton}
\label{subsec:polyptoton}
\paragraph{Definition}Polyptoton is the repetition of a word but in a different inflectional form. This figure is barely present in languages without inflections, such as English~\cite{fahnestock2002rhetorical}. The problem is that this figure is often neglected, as most research about rhetorical figures is conducted in English.
\paragraph{Related Figures} Antanaclasis, where the word is perfectly lexically repeated but with a different function (where a noun can also be a verb, e.g., sign).
\paragraph{Rhetorical Effect}Fahnestock~\cite{fahnestock2002rhetorical} mentions that polyptoton is crucial in successful argumentation as it takes a widely accepted concept and transfers it, making the relation between those two words obvious, significant, and therefore more memorable.
\paragraph{Domain}Fahnestock~\cite{fahnestock2002rhetorical} describes that polyptoton is frequently used in science, especially in chemistry and biology for nomenclature and taxonomy to express the common basis but also to highlight the distinction (e.g., alkanes, alkenes, alkynes)
\paragraph{Datasets} Gawryjolek~\cite{gawryjolek2009automated} collects instances of polyptoton. Java~\cite{java2015characterization} collected 50 instances of polyptoton.
\paragraph{Detection}

Gawryjolek~\cite{gawryjolek2009automated} first stems words, then deletes and adds prefixes to check if those forms are present in Wordnet. For most of his approaches to detect rhetorical figures that are formed by a repetition of words, he can only find exactly repeating words. If there is a change in the inflection, he is not able to find the repetition. Alliheedi and Di Marco~\cite{alliheedi2014rhetorical} investigate the detection of polyptoton, too, by using Gawryjolek's approach. However, they reveal that most polyptotons cannot not be found, without indicating how many exactly.

Java~\cite{java2015characterization} uses a very similar approach to Gawryjolek by referring to Wordnet for related or derived words and adding prefixes and suffixes.

Lawrence et al.~\cite{lawrence2017harnessing} use stemming on their data and look for the same words with different endings, which marks a change in flexion and, therefore, indicates a polyptoton.

\begin{table}[!h]
\begin{tabular}{llllrrr} \toprule
Figure                    & Paper      & Year & Approach & Precision & Recall & F1 \\
\midrule
\multirow{4}{*}{Polyptoton}
& Gawryjolek~\cite{gawryjolek2009automated} & 2009 & rule-based& - & - &  \\
& Alliheedi and Di Marco~\cite{alliheedi2014rhetorical} & 2014 & rule-based& - & - &  \\
& Java~\cite{java2015characterization} & 2015 & rule-based& 96.00 & 90.00 & \textbf{92.90} \\
& Lawrence et al.~\cite{lawrence2017harnessing} & 2017 & rule-based& - & - &  \\

\bottomrule
\end{tabular}
\caption{Overview of approaches to detect polyptoton and their performance metrics on different datasets.}
  \Description[Polyptoton approaches are all rule-based]{All approaches are rule-based. Java achieves an F1-score of 92.90 percent.}
\end{table}

\subsection{Polysyndeton}
\label{subsec:polysyndeton}
\paragraph{Definition} Polysyndeton joins words with conjunctions in between. Example: ``The mother and the father and the grandma''.
\paragraph{Related Figures} Polysyndeton is related to asyndeton, as it is its opposition. In an asyndeton, conjunctions between words are left out.
\paragraph{Rhetorical Effect}Polysyndeton slows down the rhythm, making a pause or creating a balance between the polysyndetic words. 
\paragraph{Domain} It can illustrate the lack of grammar of a character in a theater role or create enthusiasm and surprise in the audience~\cite{hamza2020polysyndeton}. However, the figure is found in every other domain where the emphasis through a pause is necessary.
\paragraph{Datasets} Gawryjolek~\cite{gawryjolek2009automated} and Java~\cite{java2015characterization} again collect examples of polysyndeton (28 sentences in the case of Java).
\paragraph{Detection}

Gawryjolek~\cite{gawryjolek2009automated} looks within two sentences for a polysyndeton to find a connecting conjunction.
Java~\cite{java2015characterization} sets a single sentence window and looks for conjunctions. He achieves a precision and recall of 100~\% while searching within a single sentence.
Lagutina et al.~\cite{lagutina2020automatic} use as input for their polysyndeton detection algorithm a pair of conjunctions and search for words that match. If this conjunction is repeated at least two times, it is considered to be a polysyndeton.

\begin{table}[!h]
\begin{tabular}{llllrrr} \toprule
Figure                    & Paper      & Year & Approach & Precision & Recall & F1 \\
\midrule
\multirow{3}{*}{Polysyndeton}
& Gawryjolek~\cite{gawryjolek2009automated} & 2009 & rule-based& - & - &  - \\

& Java~\cite{java2015characterization} & 2015 & rule-based& 100.00 & 100.00 & \textbf{100.00} \\
& Lagutina et al.~\cite{lagutina2020automatic} & 2020 & rule-based& - & - &  - \\

\bottomrule
\end{tabular}
\caption{Overview of approaches to detect polysyndeton and their performance metrics.}
\end{table}

\subsection{Rhetorical Question}
\label{subsec:rhetoricalQuestion}
\paragraph{Definition}A rhetorical question is a question that does not expect an answer and does not have the goal of obtaining information but is rather a statement~\cite{burton2007forest}.

A major challenge in detecting rhetorical questions is that they do not differ syntactically from normal questions~\cite{ranganath2018understanding}. Furthermore, prosody, pronunciation, and intonation play a major role in creating a rhetorical question~\cite{dehe2020prosody,bhatt1998argument,ranganath2018understanding}, making it more difficult to detect it in written form.

\paragraph{Related Figures} Brown and Levinson~\cite{brown1978universals} imply a co-location of rhetorical questions with the rhetorical figure sarcasm, which has not yet been addressed in the literature. The Silva Rhetoricae~\cite{burton2007forest} mentions the figure anacoenosis, which is a special form of a rhetorical question where the opinion of the audience is required but should be in line with the questioners.
\paragraph{Rhetorical Effect} Rhetorical questions engage the audience and create attention. It also forces the audience to reflect on the statement that the question implies.
\paragraph{Domain}Rhetorical questions are used in persuasion techniques to influence people~\cite{gass2022persuasion,ranganath2018understanding,anzilotti1982rhetorical}. Therefore, it is more likely to encounter them in situations where persuasion plays a major role, such as political speeches or sales talks~\cite{frank1990you}.

In a study, Petty et al.~\cite{petty1981effects} test the persuasive effect of messages with strong and weak arguments on people on topics with either no or high personal relevance. Using rhetorical questions in messages with low personal relevance but strong arguments makes the message more persuasive, whereas messages with weak arguments become less persuasive with rhetorical questions. Messages with high personal relevance and strong arguments become less persuasive, whereas the persuasiveness is increased by rhetorical questions in messages with high personal relevance and weak arguments. Ranganath et al.~\cite{ranganath2018understanding} show a significant sentiment change between previous messages of users and the following message containing a rhetorical question. Their assumption is that rhetorical questions are used to strengthen or mitigate previously expressed sentiments.
Brown and Levinson~\cite{brown1978universals} consider rhetorical questions as a ``politeness strategy'' which expresses ``excuses, criticism, or sarcastic remarks.'' Chen et al.~\cite{chen2021jointly} show a connection between rhetorical questions and the emotion of ``disgust''.
The detection of rhetorical questions would improve the performance of applications such as question answering, document summarization, author identification, information extraction, or opinion extraction~\cite{bhattasali2015automatic}.

\paragraph{Datasets}
Zhu et al.~\cite{zhu2022configure} have 1185 discourse fragments of rhetorical questions in their ConFiguRe corpus of Chinese figures.

Chen et al.~\cite{chen2021jointly} collect rhetorical questions in Chinese from literature, textbooks, microblogs, and websites. In addition, they annotate emotions and show which figure tends to be connected to which emotion.

Morio et al.~\cite{morio2019revealing} investigate persuasive argumentation in online forums. They structure the dialogues in discourse units, where one of them is a rhetorical statement expressed by ``figurative phrases, emotions, or rhetorical questions''. They have a manually annotated dataset but do not specify which figures they focus on exactly or how they differentiate between normal or rhetorical questions.

\paragraph{Detection}
Bhattasali et al.~\cite{bhattasali2015automatic} state that there is no prior work considering rhetorical questions in dialogue. In their research, they achieve the best results when including context information. They observe that certain cue words, context information, and strong negative polarity items reveal if a question is rhetorical or not. They use Naive Bayes and SVM with n-grams and POS bi- and trigrams from precedent and subsequent sentences as features. Taking into account precedent, subsequent, and features of the question achieves the best result with an F1 of 53.71~\%. However, only 5~\% of questions in their dataset are rhetorical questions.

Ranganath et al.~\cite{ranganath2016identifying} (extended version:~\cite{ranganath2018understanding}) collect Tweets from Twitter (nowadays called X) that contain question marks and one of the hashtags ``\#Rhetoricalquestion'' or ``\#Dontanswerthat''. In addition, they randomly sample questions. To ensure that they do not contain rhetorical questions, they are annotated by crowdsource workers. Interestingly, the authors find a connection between context, sentiment, and rhetorical questions. A rhetorical question is more likely to address the same topic as previous status messages. Furthermore, rhetorical questions show a shift in sentiment compared to the previous messages of users. 
The annotation process can be seen as a point of criticism in this paper, as the crowd workers only annotate if the randomly sampled questions are rhetorical. The authors rely completely on the hashtags set by Twitter users. However, their rhetorical competency should be questioned.

\begin{table}[!h]
\begin{tabular}{llllrrr} \toprule
Figure                    & Paper      & Year & Approach & Precision & Recall & F1 \\
\midrule
\multirow{2}{*}{Rhetorical Question}

& Bhattasali et al.~\cite{bhattasali2015automatic} & 2015 & model-based& - & - & 53.71  \\
& Ranganath et al.~\cite{ranganath2016identifying} & 2016 & model-based & - & - &  \textbf{69.73} \\

\bottomrule
\end{tabular}
\caption{Overview of approaches to detect rhetorical questions and their performance metrics on different datasets.}
\end{table}







\subsection{Zeugma}
\label{subsec:zeugma}
\paragraph{Definition}Zeugma is a rhetorical device in which a single word, usually a verb, governs or modifies two or more words in a sentence, linking two different lines of thoughts. Often, the use of this verb is grammatically or logically correct, with only one of them usually. This creates a form of clever or unexpected linkage between distinct elements in the sentence, often resulting in humor, irony, or a nuanced expression.
The Silva Rhetoricae~\cite{burton2007forest} differentiates between four special forms of zeugma, depending where the affected verb appears: prozeugma (beginning position),    hypozeugma (ending position), epizeugma (beginning or ending position), mesozeugma, and synzeugma (middle position). An example of a zeugma is ``he stole my heart and my wallet'', where stole is the governing verb.
\paragraph{Related Figures} Irony appears often when zeugma is used in a comical way. Furthermore, when a metaphorical non-literal meaning is combined with a literal meaning, the humorous effect increases.
\paragraph{Rhetorical Effect} creates humorous effects, surprises, and therefore increased emotional affection and memorability, e.g., ``Take my advice; I don’t use it anyway.''\cite{mihalcea2006learning}.
Tartakovsky and Yeshayahu~\cite{tartakovsky2023storm} consider zeugma as ``a breach of syntactic iconicity''. They say that zeugma implies a semantic connectedness between the two objects the verb refers to while they are, in fact, distant. Although they are nonsymmetrical in a semantic way, they are put in a syntactically symmetrical way, causing this breach that creates a surprising effect.

\paragraph{Domain} Because of the humorousf effect that zeugma creates, it is often used in comedy or comic situations.
\paragraph{Datasets}
A dataset for zeugma in Czech was built by Medkova~\cite{medkova2021building}.

\paragraph{Detection}

Medkova~\cite{medkova2020automatic} uses a rule-based approach to detect the figure zeugma in the Czech language. They collect grammar rules and words with their proposition from a lexicon. With their formulated rules, they can check if two verbs have the same object. They use the cztenten17 corpus~\cite{jakubivcek2013tenten} as dataset and manually change 17 sentences to include a zeugma. Although their dataset consists of 5313 sentences, of which 1681 contain a zeugma, they only focus on one specific form of zeugma with the conjunction ``a''. However, the author neither mentions how they annotate the instances nor does she report inter-annotator ratings. In her later work~\cite{medkova2022distinguishing}, Medkova extends the previous dataset~\cite{medkova2021building} and fine-tunes a pre-trained Czech language model called ZeugBERT. She achieves an accuracy of 88~\%.

\begin{table}[!h]
\begin{tabular}{llllrrr} \toprule
Figure                    & Paper      & Year & Approach & Precision & Recall & F1 \\
\midrule
\multirow{2}{*}{Zeugma}

& Medkova~\cite{medkova2020automatic}  & 2020& rule-based& - & - &  \\
& Medkova~\cite{medkova2022distinguishing} & 2022 & model-based& - & - &  \\

\bottomrule
\end{tabular}
\caption{Overview of approaches to detect zeugma and their performance metrics on different datasets.}
\end{table}

\section{Current Challenges and Research Gaps in Computational Approaches towards Rhetorical Figures}
\label{sec:currentChallenges}
\begin{figure}[!h]
    \centering
    \includegraphics[width=\linewidth]{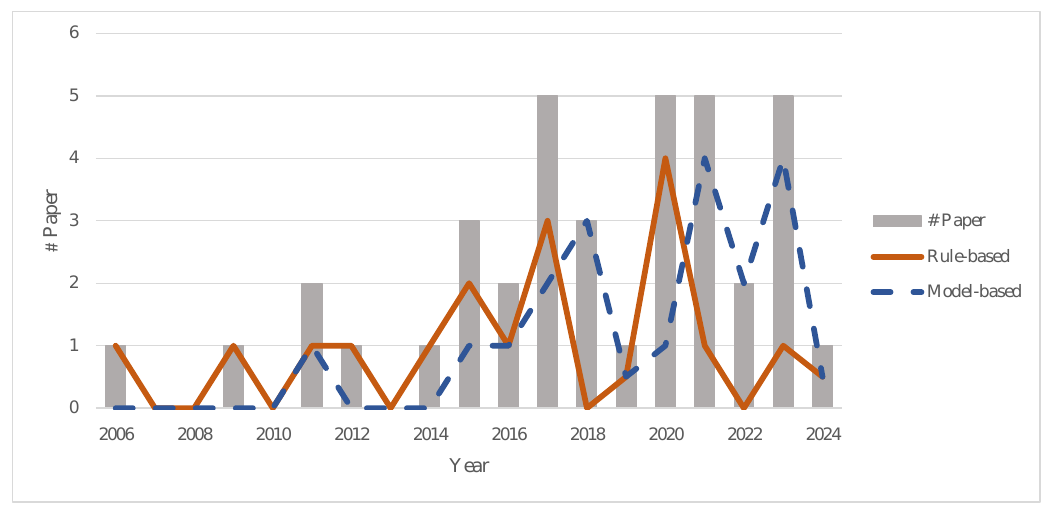}
    \caption{Distribution of papers over the years and number of rule- and model-based methods.}
    \Description[Distribution of rule-based and model-based approaches in the papers over time.]{An increasing research interest in rhetorical figure detection starting at around 2018. Rule-based approaches are decreasing while model-based approaches are increasing.}
    \label{tab:paperdist}
\end{figure}

In this survey, we investigate computational approaches to detect 26 lesser-known rhetorical figures in 39 papers. Table~\ref{tab:paperdist} shows the publication dates of the papers and the distribution between rule- and model-based approaches. 

Paida~\cite{paida2019double} in 2019 uses both rule- and model-based approaches and compares them, while Kühn et al.~\cite{kuhn2024using} use a multi-level pipeline combining a rule-based with a model-based approach. Please note that the dates for 2024 are only preliminary. The ``outlier'' of Fass~\cite{fass1991met} in 1991 is not shown for clarity. The distribution of rule- and model-based methods is balanced, with an increasing trend of model-based algorithms.
In total, the 39 papers consider 86 different approaches for multiple figures that we list in the tables at the end of each subsection. Table~\ref{tab:approach_dist} gives more insights about the approach category. Here, rule-based approaches are used for the detection of more figures, especially caused by papers investigated many figures with rule-based approaches, like Gawryjolek~\cite{gawryjolek2009automated} or Java~\cite{java2015characterization}. As Paida~\cite{paida2019double} uses both categories, we counted the approaches for both categories.

Table~\ref{tab:language_dist} shows which languages were considered in the different approaches. As Hromada~\cite{hromada2011initial} and Lagutina et al.~\cite{lagutina2019automated} considered two languages, we counted them twice.
\begin{table}[!h]
\begin{tabular}{lrr} \toprule
Approach category                    & \# Approaches      & In Percent  \\
\midrule
Rule-based & \textbf{60} & \textbf{68.97 \%} \\
Model-based & 24 & 27.59 \% \\
Rule-\& Model-based & 1 & 1.15 \% \\
Unknown & 2 & 2.30 \% \\
\bottomrule
\end{tabular}
\caption{Distribution of the approach categories over the 86 presented approaches.}
\label{tab:approach_dist}
\end{table}

\begin{table}[!h]
    \centering
    \begin{tabular}{lrr}
    \toprule
    Language  &\# Approaches & In Percent\\ \midrule
    English  & \textbf{74} & \textbf{69.81 \%} \\
    German & 10 & 9.43 \% \\
    Russian & 8 & 8.16 \% \\
    French & 4 & 7.55 \% \\
    Latin & 4 & 7.55 \% \\
    Chinese & 3 & 3.06 \% \\
    Czech & 2 & 2.83 \% \\
    Japanese & 1 & 0.94 \% \\ \bottomrule
    \end{tabular}
    \caption{Distribution of languages.}
    \label{tab:language_dist}
\end{table}

These insights reveal that \textbf{rule-based approaches} are still favored in the domain of rhetorical figures, although the number of model-based approaches has increased over recent years. 
Zhu et al.~\cite{zhu2022configure} mention that their rule-based approaches achieve lower performance results than their end-to-end methods, suggesting that the complex task of detecting rhetorical figures cannot be solved by identifying ``shallow and obvious patterns.'' Therefore, we see much potential in this domain by trying machine- or deep-learning-based approaches, especially with the advent of large language models.

Furthermore, we show once more that the NLP community is dominated by approaches and tools for the \textbf{English language}, making the detection of rhetorical figures in other languages even more challenging. 
A direct translation from another language into English to be able to use existing tools is not an option. Translation can lead to a loss of the specificities of figures, e.g., a change in the number of syllables, or losing the parallel syntax~\cite{kuhn2023hidden,kuhn2024using}. Furthermore, some figures are language dependent, i.e., a rhetorical figure in English does not have a matching counterpart in another language~\cite{zhu2022configure,kuhn2022grhoot}, or figures with the same name have deviating definitions in different languages~\cite{wang2022towards}.
Another problem with focusing on English is that it lacks grammatical gender and is a language without inflection. Some figures based on a change in inflections (such as polyptoton), therefore, occur more rarely than in languages with strong inflections (e.g., German) \cite{fahnestock2002rhetorical}.
We think that it is necessary to reward research that focuses on other languages than English. Furthermore, we are confident that the computational rhetorical figure detection in other languages could benefit from including syntactical information, e.g., syntax trees. 

We want to summarize more findings that we identified during the research for our survey. One of the biggest problems for most researchers is the \textbf{lack of data} or \textbf{imbalanced datasets}~\cite{ranganath2018understanding, bhattasali2015automatic,kuhn2023hidden,kuhn2024using,dubremetz2017machine,adewumi2021potential}. Popular figures appear relatively often in a text or tagged by users in social media posts (e.g., \texttt{\#rhetoricalquestion}\cite{ranganath2018understanding}). This makes it easier for researchers to build larger annotated datasets. Figures that also play an important role in persuasive communication but are not that frequently used are often neglected, as it is difficult to find instances of them~\cite{dubremetz2015rhetorical}. 
When creating datasets, an inherent bias poses a problem, as often only sentences are chosen with salient instances of rhetorical figures. This means that edge cases, where it is arguable whether it is a rhetorical figure or not, are not included in the dataset in the first step. 
If a dataset is imbalanced, using accuracy as a performance metric can be highly deceptive. Assume that 90~\% of data samples are text without a rhetorical figure, whereas 10~\% are examples of a certain rhetorical figure. If a classifier now always predicts the negative class for all data samples, it will achieve a classification accuracy of 90~\%. Therefore, evaluation metrics must be chosen wisely, and appropriate countermeasures have to be taken, such as appropriate augmentation techniques or over- or undersampling. Other metrics that are robust to imbalances need to be included, such as precision and recall. Nevertheless, they have to be considered carefully as their problems became obvious in Gawryjolek's~\cite{gawryjolek2009automated} and Java's work~\cite{java2015characterization}. 
Furthermore, with only a few datasets with positive examples of rhetorical figures, it is more difficult to train machine models on~\cite{dubremetz2017machine,zhang2021mover} or fine-tune language models to achieve better performance. Also, those large language models seem to be a solution to this problem. With language generation, instances of rhetorical figures can be generated. A downside is that the LLM was probably pre-trained on data in which rhetorical figures other than metaphor are not that frequent, making the generation more difficult.

Another problem we often see is the definition of the \textbf{boundaries} where to look for a figure. As figures can span over multiple sentences, paragraphs, or the whole text, it is important to define where to start and end where to look for a figure. If a repetition of two words is too far apart, it is not recognized as salient anymore by humans, while automatic parsers identify the repetition~\cite{strommer2011using}. Properly defining boundaries determines the success of rhetorical figure detection~\cite{strommer2011using}. We think that it is important for future dataset constructions to include not only one or two sentences containing the figure but also more \textbf{context} to find exactly such cases where the boundary is blurry. Context is always important to understand rhetorical figures~\cite{ranganath2018understanding,lawrence2017harnessing,troiano2018computational}.

As we showed in Section~\ref{sec:lingview}, definitions, names, and spellings of rhetorical figures are often \textbf{inconsistent}~\cite{gavidia2022cats,harris2023rules}. 
Magu et al. \cite{magu2018determining} describe the detection of euphemism but, in fact, focus more on codewords and their translation. This topic is nevertheless related to euphemism, but it is arguable if it is, in fact, a euphemism. Another example is the research of Dubremetz, in which she describes the figure chiasmus but actually refers to what is known as antimetabole~\cite{schneider2021data}. Java~\cite{java2015characterization} approaches the figure isocolon as a parallelism. As a solution, we suggest consulting different sources before approaching the figure detection task. In addition, we suggest to introduce a ranking scheme for every figure. Some researchers worked with those ranking schemes~\cite{dubremetz2015rhetorical,troiano2018computational,zhang2021mover}, but most work see rhetorical figure detection as a binary classification task. After our survey, we think figures should be categorized according to their saliency or conspicuousness. A \textbf{non-binary ranking scheme} would also help define whether a figure appears intentionally or accidentally. Especially in the case of repetition, it is questionable if it is an accidental or intentional repetition with a rhetorical purpose~\cite{strommer2011using, dubremetz2015rhetorical}. 
This leads to the problem that annotators often cannot agree if it is actually a figure and which figure it is, decreasing the agreement between annotators and the reliability of the annotation itself. Strommer~\cite{strommer2011using} describes that on his 156 instances, the annotators only agree on two of them as intentional anaphoras. Troiano et al.~\cite{troiano2018computational} also mention that they had diverse annotations in their hyperbole set. 

Another aspect we want to remark is that none of the approaches uses \textbf{rhetorical ontologies} for the detection. 
Those ontologies aim to formalize rhetorical figures and represent all the different definitions, spellings, and construction rules in a machine-readable format. The most important ontologies in this domain are 
the English RhetFig ontology~\cite{kelly2010toward}, the Serbian RetFig ontology by Mladenović and Mitrović~\cite{mladenovic}, the English Ploke ontology~\cite{wang2021ontology}, and the German GRhOOT ontology~\cite{kuhn2022grhoot}. Every mentioned ontology contains various rhetorical figures and their specificities. However, none of the investigated approaches seemed to use those helpful resources. We want to encourage researchers to include those ontologies in the future. Especially in the era of large language models, the information from those ontologies can be used for in-context learning to generate or detect rhetorical figures.

\section{Conclusion and Future Work}
\label{sec:concl}
This survey highlights the importance of rhetorical figures in the domain of NLP - their modeling, representation, detection, and inclusion in various tasks. We show how the figures can improve the performance in hate speech detection, sentiment analysis, or humor identification. This survey sheds light on detection approaches for lesser-known, yet omnipresent rhetorical figures. Furthermore, we present 86 approaches and their take on the task of rhetorical figure detection. Our key finding is that many approaches still rely on rule-based algorithms, not using the full potential of model-based algorithms or large language models. 

We hope to inspire more researchers to go on a quest of computationally detecting rhetorical figures, especially those besides the popular metaphor, irony, or sarcasm. To improve NLP systems in general and language understanding especially, it will be crucial to detect and interpret rhetorical figures correctly in the given context. Our survey provides a solid starting point by giving an overview of different rhetorical figures, their rhetorical effects, related figures with similar properties, datasets, and detection approaches.


\section{Acknowledgments}

\begin{figure}[h!]
	\centering
	\includegraphics[width=0.25\linewidth]{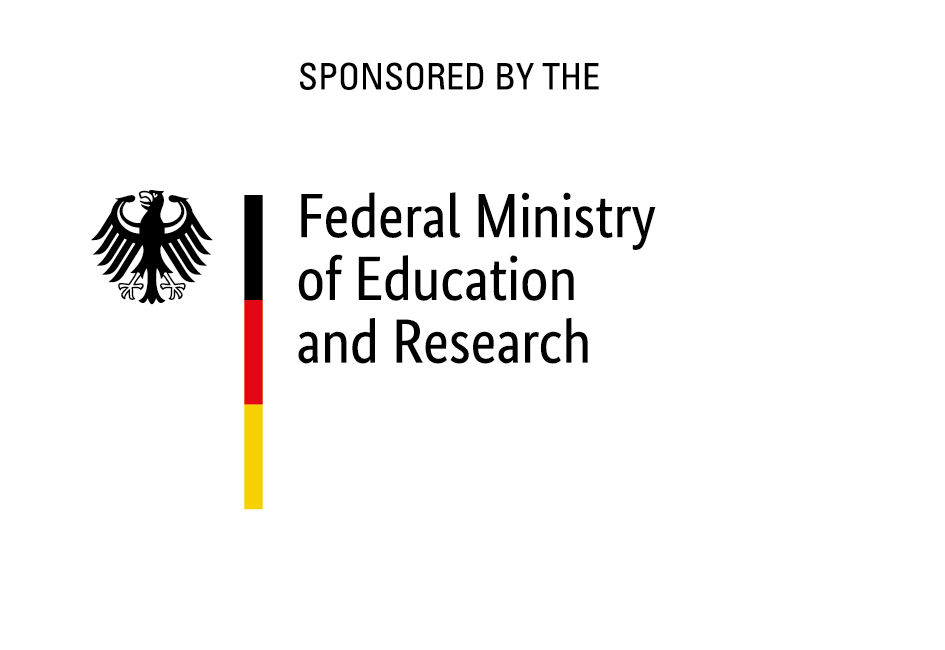}
   \Description[Logo of the German ministry of education and research]{}
\end{figure}

The project on which this report is based was funded by the German Federal Ministry of Education and Research (BMBF) under the funding code 01|S20049. The author is responsible for the content of this publication.

\bibliographystyle{ACM-Reference-Format}
\bibliography{tex/08_references.bib}

\appendix


\end{document}